\documentclass{article}

\usepackage[accepted]{icml2019}
\usepackage{float}

\usepackage{latex-preamble/preamble}
\newcommand{\vf}{\Var_{q_1}[f(x_1) w_1]}
\newcommand{\vfs}{\Var_{q_2}[f(x^*_1) w^*_1]}

\long\def\remark#1{%
    \ifvmode\else
        \unskip\raisebox{-4.5pt}[0pt][0pt]{\rlap{$\scriptstyle\diamond$}}%
    \fi
    \setlength\marginparwidth{1.5cm}
    \marginpar{\raggedright\hbadness=10000
    \parindent=8pt \parskip=2pt
    \def\baselinestretch{0.8}\tiny
    \itshape\noindent #1\par}}

\usepackage{amsthm}
\usepackage{thmtools}
\usepackage{thm-restate}
\usepackage{hyperref}
\usepackage{adjustbox}

\theoremstyle{plain}

\theoremstyle{remark}
\theoremstyle{assumption}

\icmltitlerunning{Amortized Monte Carlo Integration}

\begin{document}

\setlength{\abovedisplayskip}{5pt}
\setlength{\belowdisplayskip}{5pt}
\setlength{\abovedisplayshortskip}{5pt}
\setlength{\belowdisplayshortskip}{5pt}	
\setlength{\parskip}{5pt}

\twocolumn[
\icmltitle{Amortized Monte Carlo Integration}

\icmlsetsymbol{equal}{*}

\begin{icmlauthorlist}
\icmlauthor{Adam Goli{\'n}ski}{equal,oxstats,oxeng}
\icmlauthor{Frank Wood}{ubccs}
\icmlauthor{Tom Rainforth}{equal,oxstats}
\end{icmlauthorlist}

\icmlaffiliation{oxstats}{Department of Statistics, University of Oxford, United Kingdom}
\icmlaffiliation{oxeng}{Department of Engineering Science, University of Oxford, United Kingdom}
\icmlaffiliation{ubccs}{Department of Computer Science, University of British Columbia, Vancouver, Canada}

\icmlcorrespondingauthor{Adam Goli{\'n}ski}{adamg@robots.ox.ac.uk}

\icmlkeywords{amortized inference, Monte Carlo}

\vskip 0.3in
]

\printAffiliationsAndNotice{\icmlEqualContribution}  %

\begin{abstract}
Current approaches to amortizing Bayesian inference 
focus solely on approximating the posterior distribution.
Typically, this approximation is, in turn, used to calculate expectations 
for one or more target functions---a computational pipeline which is
inefficient when the target function(s) are known upfront. 
In this paper, we address this inefficiency by introducing AMCI,
a method for \emph{amortizing Monte Carlo integration} directly. 
AMCI operates similarly to amortized inference but produces three distinct amortized proposals, 
each tailored to a different component of the overall expectation calculation. At run-time,
samples are produced separately from each amortized proposal, before being combined to an overall estimate of the expectation.
We show that while existing approaches are fundamentally limited in the level of
accuracy they can achieve, AMCI can theoretically
produce arbitrarily small errors for any integrable target function 
using only a single sample from each proposal at run-time.  We further
show that it is able to empirically outperform the theoretically optimal
self-normalized importance sampler on a number of example problems.
Furthermore, AMCI allows not only for amortizing
over datasets but also amortizing over target functions.

\end{abstract}

\section{Introduction}
\label{sec:intro}

At its core, Bayesian modeling is rooted in the calculation of expectations:
the eventual aim of modeling is typically to make a decision or to construct
predictions for unseen data, both of which take the form of an expectation under
the posterior~\citep{Robert2007}. This aim can thus be summarized in the form of one or more
expectations $\E_{p(x|y)}\! \left[f(x)\right]$, where $f(x)$ is a target function and
$p(x|y)$ is the posterior distribution on $x$ for some data $y$, 
which we typically only know up to a normalizing constant $p(y)$. 
More generally, expectations with respect to distributions with unknown normalization constant
are ubiquitous throughout the sciences~\cite{Robert2013}.

Sometimes $f(x)$ is not known up front. Here it is typically
convenient to first approximate $p(x|y)$, e.g. in the form of 
Monte Carlo (MC) samples, and then later use this approximation to calculate estimates, rather than
addressing the target expectations directly.

However, it is often the case in practice that a particular target function, or class of
target functions, is known a priori. 
For example, in decision-based settings $f(x)$ takes the form of a loss function, while any
posterior predictive distribution constitutes a set of expectations with respect to the posterior, 
parameterized by the new input.
Though often overlooked, it is well established that in such \emph{target-aware}
settings the aforementioned pipeline of first approximating $p(x|y)$ and then using this
as a basis for calculating $\E_{p(x|y)}\! \left[f(x)\right]$ is 
suboptimal as it ignores relevant information in $f(x)$~\citep{Hesterberg1988,Wolpert1991,Oh1992,Evans1995,
	Meng1996,Chen1997,Gelman1998,Lacoste-Julien2011,Owen2013,Rainforth2018it}.
As we will later show, the potential gains in such scenarios can be arbitrarily large.

In this paper, we extend these ideas to
\emph{amortized} inference settings \citep{Stuhlmuller2013, Kingma2014, Ritchie2016, Paige2016, Le2016, Le2017,
	Webb2018}, wherein one looks to amortize the cost of inference
across different possible datasets
by learning an artifact that assists the inference process
at runtime for a given dataset.
Existing approaches do not operate in a target-aware fashion, such that even if 
the inference network learns proposals that perfectly match
the true posterior for every possible dataset, the resulting
estimator is still sub-optimal. 

To address this, we introduce AMCI, a
framework for performing \emph{Amortized Monte Carlo Integration}. 
Though still based on learning amortized proposals distributions,
AMCI varies from standard amortized inference approaches in three respects.
First, it operates in a target-aware fashion, incorporating
information about $f(x)$ into the amortization artifacts. Second, rather than
using self-normalization, AMCI employs three distinct proposals for separately estimating
$\E_{p(x)}\left[p(y|x)\max(f(x),0)\right]$, $\E_{p(x)}\left[-p(y|x)\min(f(x),0)\right]$, and
$\E_{p(x)}\left[p(y|x)\right]$, before combining these into an overall estimate. 
This breakdown allows for arbitrary performance improvements compared to self-normalized importance sampling (SNIS).
Finally, to account for cases in which multiple possible target functions may be
of interest, AMCI also allows for amortization over parametrized 
functions $f(x; \theta)$.

Remarkably, AMCI is able to achieve an arbitrarily low error at run-time using only a single sample from
each proposal given sufficiently powerful amortization artifacts, 
contrary to the fundamental limitations on the accuracy of conventional amortization approaches.
This ability is based around its novel breakdown of the target expectation into separate components,
the subsequent utility of which extends beyond the amortized setting we consider here.

\section{Background}
\label{sec:background}

\subsection{Importance Sampling}
\label{sec:inf:importance}

Importance Sampling (IS), in its most basic form, is a method for approximating an expectation $\E_{\pi(x)} \left[f(x)\right]$
when it is either not possible to sample from $\pi(x)$ directly, or when the simple MC estimate, $\frac{1}{N} \sum_{n=1}^{N}
f(x_n)$ where $x_n\sim\pi(x)$, has problematically high variance~\citep{Hesterberg1988,Wolpert1991}.
Given a proposal $q(x)$ from which we can sample and for which we can evaluate the data, 
it forms the following estimate
\begin{align}
	\mu &:= 
	\E_{\pi(x)} \left[f(x)\right] 
	=\int f(x) \frac{\pi(x)}{q(x)} q(x) dx \\
	&\approx 
	\hat{\mu} := 
	\frac{1}{N}\sum\nolimits_{n=1}^{N} f(x_n) w_n 
	\label{eq:inf:importance}
\end{align}
where $x_n \! \sim \! q(x)$
and 
$w_n \! := \! \pi(x_n)\!/ \! q(x_n)$
is known as the importance weight of sample $x_n$.

In practice, one often does not have access to the normalized form of $\pi(x)$.
For example, in Bayesian inference settings, we typically have $\pi(x)=p(x|y)\propto p(x,y)$.
Here we can use our samples to both approximate
the normalization constant and the unnormalized integral.
Thus if $\pi(x)\propto \gamma(x)$, we have
\begin{align}
\E_{\pi(x)}\![f(x)]
\!=\!
\frac{\int \! \frac{f(x)\gamma(x)}{q(x)} q(x) \text{d}x}
{\int \! \frac{\gamma(x)}{q(x)} q(x) \text{d}x} 
\!\approx\!
\frac{\sum_{n=1}^{N} \! f(x_n) w_n}
{\sum_{n=1}^{N} w_n}
\label{eq:SNIS}
\hspace{-.5em}
\end{align}
where $x_n \sim q(x)$, and $w_n := \gamma(x_n)/q(x_n)$. This approach is known as
self-normalized importance sampling (SNIS). Conveniently, we can also construct the SNIS estimate
lazily by calculating the empirical measure, i.e. storing weighted samples,
\begin{align}
\pi(x) \approx \sum\nolimits_{n=1}^{N} \! w_n \delta_{x_n}(x) \Big/
\sum\nolimits_{n=1}^{N} w_n
\end{align}
and then using this to construct the estimate in~\eqref{eq:SNIS} when $f(x)$ becomes available.
As such, we can also think of SNIS as a method for Bayesian inference as, informally speaking, the
empirical measure produced can be thought of as an approximation of the posterior.

For a general unknown target, the optimal proposal, 
i.e. the proposal which results in estimator having lowest possible variance,
is the target distribution 
$q(x) = \pi(x)$ (see e.g.~\citep[5.3.2.2]{Rainforth2017}).
However, this no longer holds if we have some information about $f(x)$.
In this target-aware scenario, the optimal behavior turns out to depend on whether
we are self-normalizing or not.

For the non-self-normalized case,
the optimal proposal can be shown
to be $q^*(x) \!\propto\! \pi(x) |f(x)|$~\citep{Owen2013}.
Interestingly, in the case where $f(x) \ge 0 \,\, \forall x$, this leads to an exact estimator, i.e. 
$\hat{\mu}=\mu$ (with $\hat{\mu}$ as per~\eqref{eq:inf:importance}).
To see this, notice that the normalizing constant for $q^*(x)$ is 
$\int \pi(x) f(x) \diff x \!=\! \mu$ 
and hence
$q^*(x) \!=\! \pi(x) f(x) / \mu$.  Therefore, even when $N\!=\!1$,
any possible value of the resulting sample $x_1$ yields an $\hat{\mu}$ satisfying
$\hat{\mu} \!=\! f(x_1) \pi(x_1)/q^*(x_1) \!=\! \mu$.

In the self-normalized case, the optimal proposal instead transpires to be
$q^*(x) \!\propto\! \pi(x)|f(x)\!-\!\mu|$~\citep{Hesterberg1988}. 
In this case, one can no longer achieve 
a zero variance estimator for finite $N$ and nonconstant $f(x)$.
Instead, the achievable error is lower bounded by~\citep{Owen2013}
\begin{align}
	\label{eq:achievable}
\E [(\hat{\mu}-\mu)^2] \ge \frac{1}{N}\left(\E_{\pi(x)} [|f(x) - \mu|]\right)^2,
\end{align}
creating a fundamental limit on the performance of SNIS, even
when information about $f(x)$ is incorporated.

Given that these optimal proposals make use of the true expectation $\mu$, we will clearly not have access
to them in practice. However, they provide a guide for the desirable properties of a proposal and can be used
as targets for adaptive 
IS
methods (see \cite{Bugallo2017} for a 
recent 
review).

\subsection{Inference Amortization}
\label{sec:inference-amortization}

Inference amortization involves
learning an \emph{amortization artifact} that takes in datasets
and produces proposals tailored to the corresponding inference problems. 
This amortization artifact typically takes the form of a parametrized proposal, 
$q(x ; \varphi(y;\eta))$, which takes
in data $y$ and produces proposal parameters using an \emph{inference network} $\varphi(y;\eta)$,
which itself has parameters $\eta$.
When clear from the context, we will use the shorthand $q(x ; y, \eta)$ for this proposal.

Though the exact process varies with context,
the inference network is usually trained either by drawing latent-data
sample pairs from the joint
$p(x,y)$~\citep{Paige2016,Le2016,Le2018}, or by drawing 
mini-batches from a large dataset using stochastic variational inference 
approaches~\citep{Hoffman2013,Kingma2014,Rezende2014,Ritchie2016}.
Once trained, it provides an efficient means of approximately
sampling from the posterior of a particular dataset, 
e.g. using SNIS.

Out of several variants, we focus on 
the method introduced by
\citet{Paige2016}, as this is the one AMCI builds upon.
In their approach, $\eta$ is trained to minimize the expectation of
$D_{KL} \left[ p(x|y) \mid\mid q(x;y, \eta) \right]$
across possible datasets $y$, giving the objective
\begin{flalign}
\mathcal{J}(\eta)
&= \mathbb{E}_{p(y)}\left[D_{KL} \left[ p(x|y) \mid\mid q(x;y, \eta) \right] \right] \nonumber \\[3pt]
&=
\mathbb{E}_{p(x,y)}\left[- \log q(x ;y, \eta)\right] 
+ \text{const wrt }\eta
\label{eq:inf-comp:original}
\end{flalign}
We note that the distribution $p(y)$ over which we are taking the expectation is actually chosen somewhat arbitrarily:
it simply dictates how much the network prioritizes a good amortization for one dataset over another; different choices
are equally valid and imply different loss functions.

This objective requires us to be able to sample from 
the joint distribution $p(x,y)$ and it can be optimized using gradient methods 
since the gradient can be easily evaluated:
\begin{flalign}
\nabla_\eta \mathcal{J}(\eta)
= \mathbb{E}_{p(x,y)}\left[- \nabla_\eta \log q(x ; y, \eta)\right].
\label{eq:inf-comp:original:gradient}
\end{flalign}

\section{AMCI}
\label{sec:amci}

Amortized Monte Carlo integration (AMCI) is a framework for amortizing
the cost of calculating expectations $\mu(y,\theta):=\E_{\pi(x;y)}[f(x;\theta)]$. Here $y$ represents changeable
aspects of the reference distribution $\pi(x;y)$ (e.g. the dataset)
and $\theta$ represents changeable parameters of the target function $f(x;\theta)$. 
The reference distribution is typically known only up to a
normalization constant, i.e. $\pi(x;y)=\gamma(x;y)/Z$ where $\gamma(x;y)$ can be evaluated pointwise, but $Z$ is unknown.
AMCI can still be useful in settings where $Z$ is known, but here we can simply use its known
value rather than constructing a separate estimator.

Amortization can be performed across $y$ and/or $\theta$. When amortizing over $y$, 
the function does not need to be explicitly parameterized; we just need to be able to 
evaluate it pointwise. Similarly, when amortizing
over $\theta$, the reference distribution can be fixed.
In fact, AMCI can be used for a parameterized set of conventional integration problems 
$\int_{x\in\mathcal{X}} f(x;\theta)dx$ by exploiting the fact that
\begin{align}
\int_{x\in\mathcal{X}} f(x;\theta)dx = \E_{\pi(x)}[f(x;\theta)/\pi(x)]
\end{align}
for any $\pi(x)$ where $\pi(x)\neq0\, \forall x \in \mathcal{X}$ for which $f(x)\neq0$.	

For consistency of notation with the amortized inference literature, we will presume a Bayesian
setting in the rest of this section, i.e. $\pi(x;y)\!=\!p(x|y)$ and $\gamma(x;y)\!=\!p(x,y)$.

\subsection{Estimator}
Existing amortized inference methods implicitly evaluate expectations 
using SNIS (or some other form of self-normalized estimator~\cite{Paige2016,Le2017}),
targeting the posterior as the optimal proposal $q^*(x;y) \approx p(x|y)$.
Not only is this proposal suboptimal when information about the target function is available,
there is a lower bound on the accuracy the SNIS approach itself can achieve as 
shown in~\eqref{eq:achievable}.

AMCI overcomes these limitations by breaking down the overall expectation into separate
components and constructing separate estimates for each.
We can first break down the target expectation into the ratio of the ``unnormalized expectation'' and
the normalization constant:
\begin{align}
\mu(y,\theta) 
& := 
\E_{p(x|y)} \!\left[f(x;\theta)\right]
=
\frac{ \E_{p(x|y)} \! \left[ f(x;\theta) \, p(y) \right]}{\E_{p(x)}\!\!\left[p(y|x)\right]}
\nonumber \\
& \phantom{:}=
\frac{\E_{q_1(x;y,\theta)}\!\!\left[\frac{f(x;\theta)p(x,y)}{q_1(x;y,\theta)}\right]}{\E_{q_2(x;y)}\!\!\left[\frac{p(x,y)}{q_2(x;y)}\right]}
=: 
\frac{E_1}{E_2}
\label{eq:estimator-for}
\end{align}
where $q_1(x;y,\theta)$ and $q_2(x;y)$ are two separate proposals,
used respectively for each of the two expectations $E_1$ and $E_2$.
We note that the proposal $q_1(x;y,\theta)$ may depend not only on the observed dataset $y$, but also on
the parameters of the target function $\theta$.

We can now generate separate MC estimates for $E_1$ and $E_2$, and take their
ratio to estimate the overall expectation:
\begin{align}
\begin{split}
&\mu(y,\theta) \approx \hat{\mu}(y,\theta) := \hat{E}_1/\hat{E}_2 \quad \text{where} \\
&\hat{E}_1 := \frac{1}{N} \sum_{n=1}^N \frac{f(x_n'; \theta) p(x_n', y)}{q_1(x_n';y,\theta)}
\quad x_n' \sim q_1(x;y,\theta) \\
&\hat{E}_2 := \frac{1}{M} \sum_{m=1}^M \frac{p(x_m, y)}{q_2(x_m;y)}
\quad x_m \sim q_2(x;y).
\end{split}
\label{eq:amci-mc-estimator}
\end{align}

The key idea behind AMCI is that we can now \textbf{separately train each of these proposals to be good
estimators for their respective expectation}, rather than rely on a single proposal to estimate both, 
as is implicitly the case for SNIS.

Consider, for example, the case where 
$f(x;\theta)\!\ge\!0$. 
If 
$q_1(x;y,\theta) \! \propto \! f(x;\theta)p(x|y)$ %
and 
$q_2(x;y) \propto p(x|y)$ %
then both $\hat{E}_1$ and $\hat{E}_2$ will form exact
estimators (as per Section~\ref{sec:inf:importance}), even if $N\!=\!M\!=\!1$.
Consequently, we achieve an exact estimator for $\mu(y,\theta)$, allowing for arbitrarily large
improvements over any SNIS estimator, because SNIS forces $q_1(x;y,\theta)$ and $q_2(x;y)$ 
to be the same distribution.

More generally, the optimal proposal for $E_1$ and $E_2$ are $q_1(x;y,\theta) \! \propto \! |f(x;\theta)|p(x|y)$ 
and $q_2(x;y) \! \propto \! p(x|y)$ respectively, with the latter always resulting in an exact
estimator for $E_2$. Thus the more $|f(x;\theta)|p(x|y)$ varies from $p(x|y)$, the worse the conventional
approach of only amortizing the posterior will perform, while the harder it becomes to construct a
reasonable SNIS estimator even when information about $f(x;\theta)$ is incorporated.
Separately learning $q_1(x;y,\theta)$ and $q_2(x;y)$ means that each will become a better individual
proposal and the overall estimator improves.

It turns out that we do not actually require the previous assumption of $f(x;\theta)\ge0 \,\, \forall x, \theta$
to achieve a zero variance estimator. Specifically, if we let\footnote{Practically, it may sometimes be beneficial
	to truncate the proposal about another point, $c$, by instead using $f^+(x;\theta)=\max(f(x;\theta)-c,0)$
	and $f^-(x;\theta)=-\min(f(x;\theta)-c,0)$, then adding $c$ onto our final estimate.}
\begin{align}
f^+(x;\theta) &= \max(f(x;\theta),0) \quad \text{and} \\
f^-(x;\theta) &= -\min(f(x;\theta),0) 
\end{align}
denote truncations of the target function into its positive and negative components (as per
the concept of posivitisation~\citep[9.13]{Owen2013}), then we
can break down the overall expectation as follows
\begin{align}
	&\mu(y,\theta) \nonumber \\
	&= 
	\frac{\E_{q_1^+(x;y,\theta)}\left[\frac{f^+(x;\theta)p(x,y)}{q_1^+(x;y,\theta)}\right]
		- \E_{q_1^-(x;y,\theta)}\left[\frac{f^-(x;\theta)p(x,y)}{q_1^-(x;y,\theta)}\right]
		}{\E_{q_2(x;y)}\left[\frac{p(x,y)}{q_2(x;y)}\right]} \nonumber
	\\
	&=: \frac{E_1^+ - E_1^-}{E_2}
	\label{eq:full-breakdown}
\end{align}
where we now have three expectations and three proposals. Analogously to~\eqref{eq:amci-mc-estimator},
we can construct estimates for each expectation separately and then combine them:
\begin{align}
&\hspace{-5pt}\mu(y,\theta) \approx \hat{\mu}(y,\theta) := (\hat{E}_1^+ - \hat{E}_1^-)/\hat{E}_2 \quad \text{where} \nonumber\\
&\hat{E}_1^+ := \frac{1}{N} \sum_{n=1}^N \frac{f^+(x_n^+; \theta) p(x_n^+, y)}{q_1^+(x_n^+;y,\theta)}
\quad x_n^+ \sim q_1^+(x;y,\theta)\nonumber \\
&\hat{E}_1^- := \frac{1}{K} \sum_{k=1}^K \frac{f^-(x_k^-; \theta) p(x_k^-, y)}{q_1^-(x_k^-;y,\theta)}
\quad x_k^- \sim q_1^-(x;y,\theta) \nonumber\\
&\hat{E}_2 := \frac{1}{M} \sum_{m=1}^M \frac{p(x_m, y)}{q_2(x_m;y)}
\quad x_m \sim q_2(x;y),
\label{eq:estimator-full}
\end{align}
which forms the AMCI estimator.
The theoretical capability of this estimator is summarized in the following result, the
proof for which is given in Appendix~\ref{sec:proof}.
\begin{restatable}{theorem}{amciThe}
\label{theorem}
If the following hold for a given $\theta$ and $y$,
\begin{align}
&\E_{p(x)} \left[f^+(x;\theta)p(y|x)\right] < \infty \\
&\E_{p(x)} \left[f^-(x;\theta)p(y|x)\right] < \infty \\
&\E_{p(x)} \left[p(y|x)\right] < \infty
\end{align}
and we use the corresponding set of optimal proposals 
$q_1^+(x;y,\theta)\propto f^+(x;\theta)p(x,y)$,
$q_1^-(x;y,\theta) \propto f^-(x;\theta)p(x,y)$, and
$q_2(x;y)\propto p(x,y)$,
then the AMCI estimator defined in~\eqref{eq:estimator-full} 
satisfies
\begin{align}
\E \left[\hat{\mu}(y,\theta)\right]&=\mu(y,\theta),
\,\,\,
\textnormal{Var}\left[\hat{\mu}(y,\theta)\right]=0
\end{align}
 for any $N\ge1$, $K\ge1$, and $M\ge1$,
such that it forms an exact estimator for that $\theta, y$ pair.
\end{restatable}
Though our primary motivation for developing the AMCI estimator is its attractive properties
in an amortization setting, we note that it may still be of use in
static expectation calculation settings. Namely, the fact that it can achieve an arbitrarily low
mean squared error for a given number of samples means it forms an 
attractive alternative to SNIS more generally, particularly when we are well-placed to
hand-craft highly effective proposals and in adaptive importance sampling settings.

We note that individual elements of this estimator have previously appeared in the literature.
For example, the general concept of using multiple proposals has been established 
in the context of multiple importance sampling~\citep{Veach1995}.
The use of two separate proposals for the unnormalized target and the normalizing constant (i.e.~\eqref{eq:amci-mc-estimator}), on the other hand, was
recently independently suggested by~\citet{Lamberti2018} in a non-amortized setting.
However, we believe that the complete form of the AMCI estimator in~\eqref{eq:estimator-full}
has not previously been suggested, nor its theoretical benefits or amortization considered.

\subsection{Amortization}
\label{sec:amci-amortization}
To evaluate~\eqref{eq:estimator-full},
we need to learn three amortized proposals $q_1^+(x;y,\theta)$, $q_1^-(x;y,\theta)$, and $q_2(x;y)$.

Learning $q_2(x;y)$ is equivalent to the standard inference amortization problem and
so we will just use the objective given by~\eqref{eq:inf-comp:original},
as described in section~\ref{sec:inference-amortization}.

The approaches for learning $q_1^+(x;y,\theta)$ and $q_1^-(x;y,\theta)$ are equivalent, other than
the function that is used in the estimators. Therefore, for simplicity, we introduce our amortization
procedure in the case where $f(x;\theta)\ge0 \,\, \forall x, \theta$, such that we can
need only learn a single proposal, $q_1(x;y,\theta)$, for the numerator as per~\eqref{eq:amci-mc-estimator}.
This trivially extends to the full AMCI setup by separately repeating the same training procedure for 
$q_1^+(x;y,\theta)$ and $q_1^-(x;y,\theta)$.

\subsubsection{Fixed function $f(x)$}
\label{sec:fixed-function}
We first consider the scenario where $f(x)$ is fixed 
(i.e. we are not amortizing over function parameters $\theta$) 
and hence in this section we drop the dependence of $q_1$ on $\theta$.

To learn the parameters $\eta$ for the
first amortized proposal $q_1(x;y,\eta)$, we need to adjust the target 
in \eqref{eq:inf-comp:original} 
to incorporate the effect of the target function.
Let $E_1(y) \!:=\! \E_{p(x)} \!\left[f(x)p(y|x)\right]$ and
$g(x|y) \!:=\! \frac{f(x) \, p(x,y)}{E_1(y)}$, 
i.e.\ the normalized optimal proposal for $q_1$.
Naively adjusting \eqref{eq:inf-comp:original} leads to a double intractable objective
\begin{flalign}
\label{eq:desired-objective}
\mathcal{J}_1'(\eta) &= \E_{p(y)} [ \KL \left( g(x|y) || q_1(x;y,\eta) \right) ] 
\nonumber \\
\begin{split}
=& \E_{p(y)} \left[ - \int_{\mathcal{X}} \frac{f(x)\,p(x,y)}{E_1(y)}\log q_1(x;y,\eta) \d x \right] \\
&+ \text{const wrt } \eta.
\end{split}
\end{flalign}
Here the double intractability comes from the fact we do not know $E_1(y)$ and, at least at the 
beginning of the training process, we cannot estimate it efficiently either.

To address this, we use our previous observation that the expectation over 
$p(y)$ in the above objective is chosen somewhat arbitrarily.  Namely, it dictates the
relative priority of different datasets $y$ during training and not the optimal
proposal for each individual datapoint; disregarding the finite capacity of the
network, the global optimum is still always $\KL \left[ g(x|y) \,||\, q_1(x;y, \eta) \right] \! = \! 0, \, \forall y$.
We thus maintain a well-defined 
objective
if we choose a different reference distribution
over datasets. 
In particular, if we take the expectation with respect to $h(y) \!\propto \! p(y) E_1(y)$, we get 
\begin{align}
\mathcal{J}_1(\eta)
&=
\mathbb{E}_{h(y)}\left[\KL \left( g(x|y) \,||\, q_1(x;y,\eta) \right)\right] 
\nonumber \\
\begin{split}
&= 
c^{-1} \,
\mathbb{E}_{p(x,y)}\left[-f(x) \log q_1(x;y,\eta)\right] \\
&\phantom{=} \, + \text{const wrt } \eta 
\label{eq:q1-fixed-f-target}
\end{split}
\end{align}
where $c = \E_{p(y)} \left[E_1(y)\right]>0$ is a positive constant that does not affect the 
optimization---it is the normalization constant for the distribution $h(y)$---and can thus be ignored.
Each term in this expectation can now be evaluated directly, meaning we can again run stochastic gradient
descent algorithms to optimize it.
Note that this does not require evaluation of the density $p(x,y)$, only the ability to draw samples.

Interestingly, this choice of $h(y)$ can be interpreted as giving larger importance to the values of $y$
which yield larger $E_1(y)$.
Informally, we could think about this choice as attempting to minimizing the L1 errors of our estimates, that is
$\E_{p(y)}[|E_1(y) - \hat{E}_1(y)|]$, presuming that the error in our estimation scales as the magnitude of the
true value $E_1(y)$.

More generally, if we choose $h(y)\!\propto \! p(y) E_1(y) \lambda(y)$ for some positive evaluable function $\lambda : \mathcal{Y} \to \mathbb{R}^+$, we
get a tractable objective of the form 
\[
\mathcal{J}_1(\eta;\lambda) = \mathbb{E}_{p(x,y)}\left[-\frac{f(x)}{\lambda(y)} \log q_1(x;y,\eta)\right]
\]
up to a constant scaling factor and offset. We can thus use this trick to adjust the relative preference given
to different datasets, while ensuring the objective is tractable.

\subsubsection{Parameterized function $f(x; \theta)$}

As previously mentioned, AMCI also allows for amortization over parametrized functions,
to account for cases in which multiple possible target functions may be
of interest. We can incorporate this by using \emph{pseudo prior} $p(\theta)$
to generate example parameters during our training.

Analogously to $h(y)$, the choice of $p(\theta)$ determines how much importance we assign to different possible functions that we would like to amortize over.
Since, in practice, perfect performance is unattainable over the entire space of $\theta$, the choice of $p(\theta)$ is important and it will have an important effect on the performance of the system. 

Incorporating $p(\theta)$ is straightforward: we take the expectation of the fixed target
function training objective over $\theta$. 
In this setting, our inference network $\varphi$ needs to take $\theta$ as input when determining the parameters of $q_1$
and hence we let
$q_1(x ; y, \theta, \eta) \!:=\! q_1(x ;\varphi(y, \theta; \eta))$.
If
$E_1(y,\theta) \!:=\! \E_{p(x)} \left[f(x;\theta)p(y|x)\right]$,
$g(x|y; \theta) \!:=\! f(x; \theta) \, p(x,y)/E_1(y, \theta)$, and
$h(y, \theta) \! \propto \! p(y) p(\theta) E_1(y, \theta)$, we get an objective which is 
analogous to~\eqref{eq:q1-fixed-f-target}:
\begin{flalign}
\label{eq:train-prop-standard}
\mathcal{J}_1(\eta) &= \mathbb{E}_{h(y, \theta)} \! \left[\KL \left( g(x|y ;\theta) \,||\, q_1(x;y, \theta, \eta) \right) \right] 
\nonumber \\
\begin{split}
=& c^{-1} \cdot
\mathbb{E}_{p(x,y) p(\theta)}\left[-f(x; \theta) \log q_1(x;y, \theta, \eta)\right] \\ 
& + \text{const wrt }\eta
\end{split}
\end{flalign}
where $c \!=\! \E_{p(y)p(\theta)} \left[E_1(y,\theta)\right] \!>\! 0$ is 
again a positive constant that does not affect the optimization.

\subsection{Efficient Training}
\label{sec:efficient-training}
If $f(x; \theta)$ and $p(x)p(\theta)$ are mismatched, i.e.~$f(x; \theta)$ is large in regions where $p(x)p(\theta)$ is low,
training by na\"{i}vely sampling from $p(x)p(\theta)$ can be inefficient. Instead, it is preferable
to try and sample from $g(\theta,x) \! \propto \! p(x)p(\theta)\!f(x;\theta)$. Though this is itself an 
intractable distribution, it represents a standard, rather than an amortized, inference problem and so it is much
more manageable than the overall training. Namely, as the samples do not depend on the proposal we are
learning or the datasets, we can carry out this inference process as a pre-training step that is
substantially less costly than the problem of training the inference networks itself.

One 
approach is to construct an MCMC sampler
targeting $g(\theta,x)$ to generate the 
samples, 
which can be done upfront before training. 
Another is to use an importance sampler
\begin{align}
\label{eq:train-prop}
&\mathcal{J}_1(\eta) = \text{const wrt }\eta \\
&+c^{-1}\mathbb{E}_{q^\prime(\theta,x)p(y|x)} \! \left[- \frac{p(\theta)p(x)f(x; \theta) }{q^\prime(\theta,x)} \log q_1(x ;y, \theta, \eta)\right]
\nonumber 
\end{align}
where $q^\prime(\theta,x)$ is a proposal as close to $g(\theta,x)$ as possible.

In the case of non-parameterized functions $f(x)$, there is no need to take an expectation over $p(\theta)$,
and we instead desire to sample from $g(x)\propto p(x)f(x)$.

\begin{figure*}[t]
	\centering
	\vspace{-5pt}
	\begin{subfigure}{\columnwidth}
		\centering
		\includegraphics[width=0.95\columnwidth]{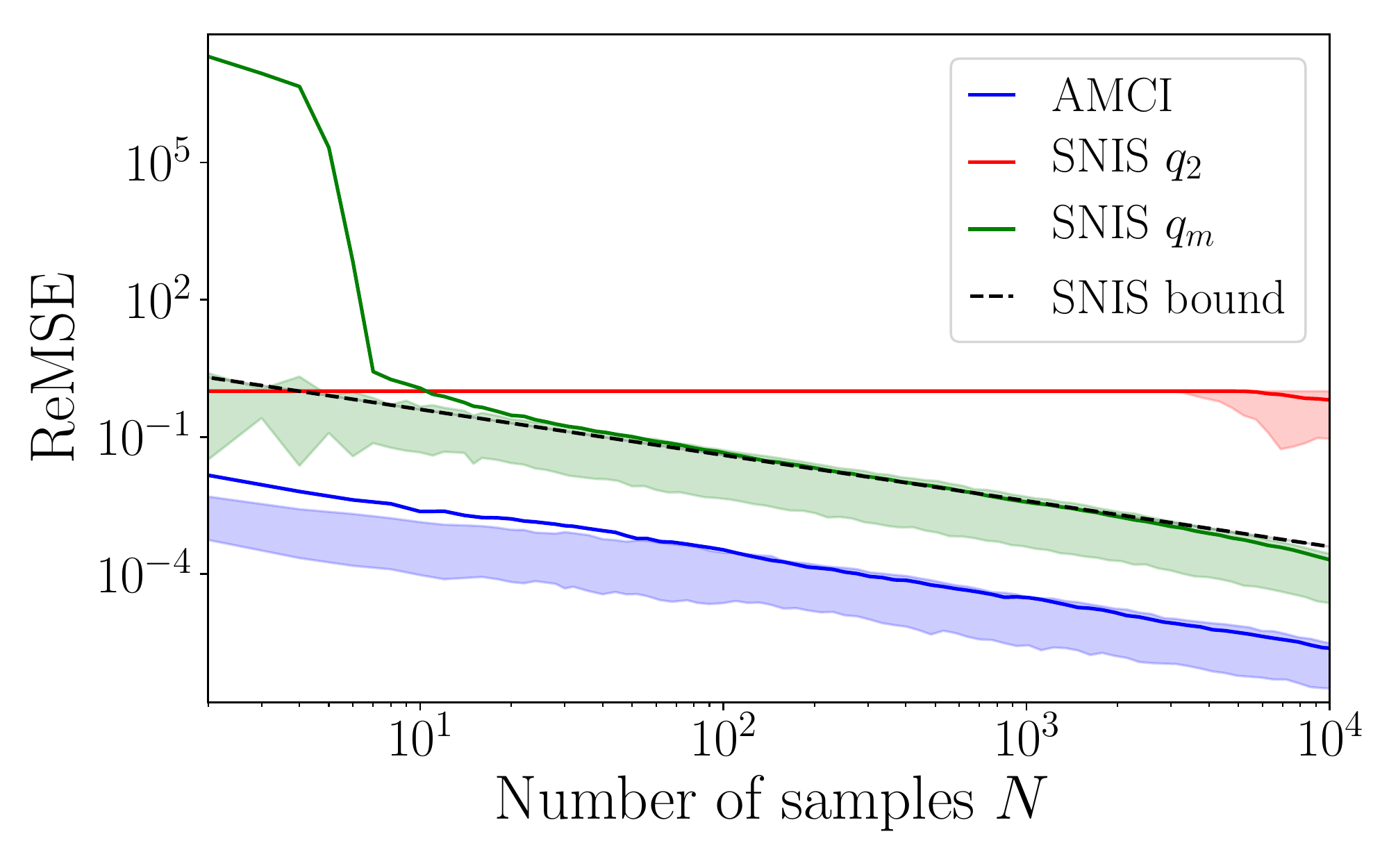}
		\vspace{-5pt}
		\subcaption{One-dimensional tail integral}
		\label{fig:one-dim}
	\end{subfigure}
	\begin{subfigure}{\columnwidth}
		\centering
		\includegraphics[width=0.95\columnwidth]{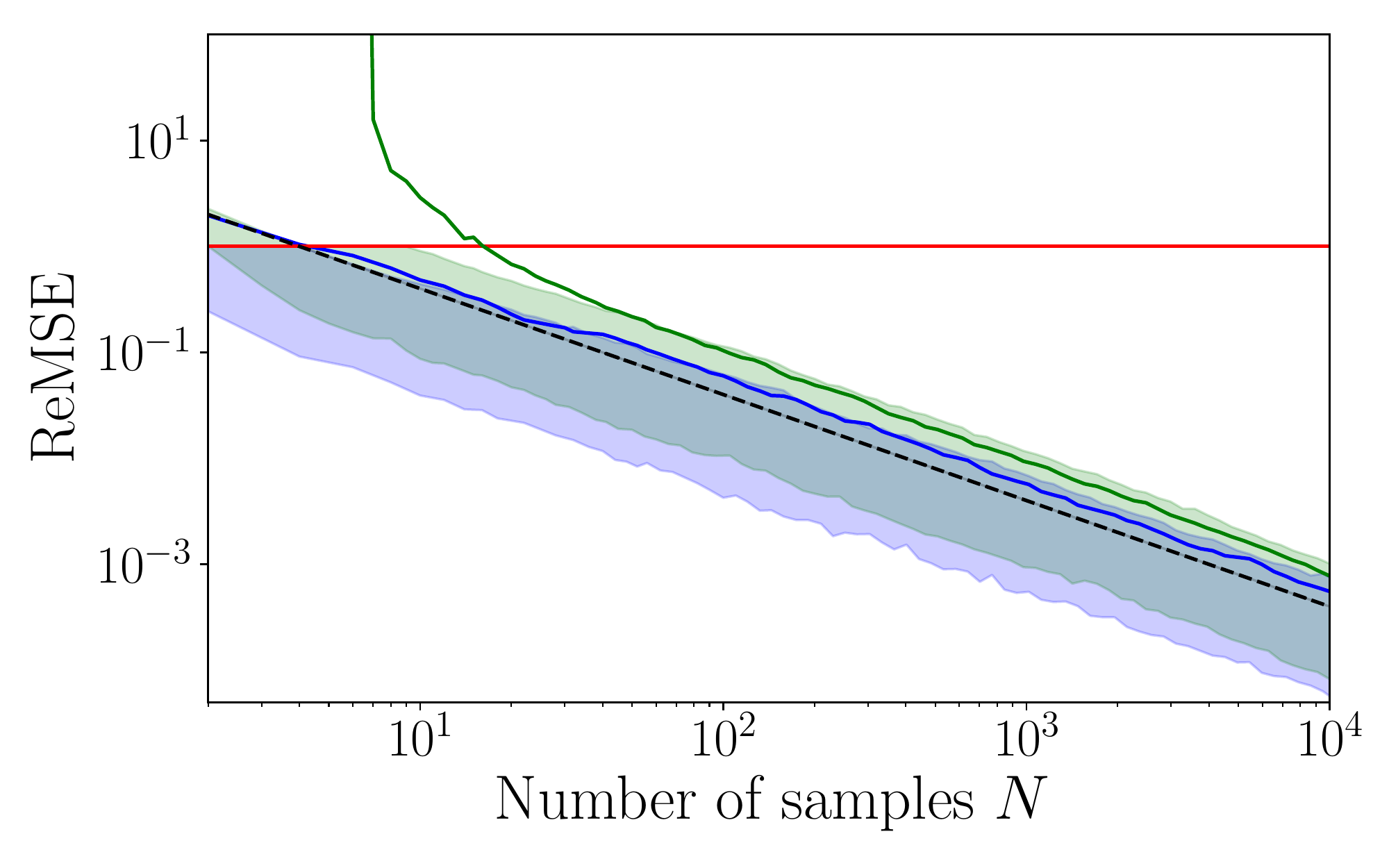}
		\vspace{-5pt}
		\subcaption{Five-dimensional tail integral}
		\label{fig:five-dim}
	\end{subfigure}
	\vspace{-5pt}
	\caption{
		Relative mean squared errors (as per~\eqref{eq:rel-error}) for [left] the one-dimensional and [right] the five-dimensional tail integral example.
		The solid lines for each estimator indicate the median of $\delta(y,\theta)$ estimated using a
		common set of $100$ samples from $y,\theta \! \sim \! p(y)p(\theta)$, with the corresponding $\delta(y,\theta)$
		then each separately estimated using $100$ samples of the respective $\hat{\delta}(y, \theta)$.  
		The shading instead shows the estimates from
		replacing $\delta(y,\theta)$ with the $25\%$ and $75\%$ quantiles of $\hat{\delta}(y, \theta)$ for a given
		$y$ and $\theta$. 
		The median of $\delta(y,\theta)$ is at times outside of this shaded region as $\delta(y,\theta)$ is often dominated by a few large outliers.
		The dashed line shows the median of $\delta(y,\theta)$ with the $\delta(y,\theta)$ corresponding to the ReMSE
		optimal SNIS estimator, namely $(\E_{p(x|y)} [|f(x;\theta)-\mu(y,\theta)|])^2/N$ as per~\eqref{eq:achievable}, which is itself estimated (with only nominal error) using $10^6$ samples.
		We note that the error for SNIS with $q_2$ proposal is to a large extent flat because there is not a single sample in the estimator for which $f(x;\theta)\!>\!0$,
		such that they return $\hat{\mu}(y,\theta)\!=\!0$ and hence give $\delta(y,\theta)\!=\!1$.
				In Figure (b) the SNIS $q_m$ line reaches the ReMSE value of $10^{18}$ at $N\!\!=\!\!2$ and the y-axis limits have been readjusted to allow clear comparison at higher $N$. 
				This effect is caused by the bias of SNIS: these extremely high errors for SNIS $q_m$ arise when all $N$ samples happen  to be drawn from distribution $q_1$, 
				for further explanation and the full picture see Figure~\ref{fig:tail5d_q1} in Appendix~\ref{sec:snisq1}.
	}
	\vspace{-10pt}
	\label{fig:tails}
\end{figure*}

\section{Experiments}
\label{sec:experiment}

Even though AMCI is theoretically able to achieve exact estimators with a finite number of samples, this will rarely be the case for practical problems, for which learning perfect proposals is not typically realistic, particularly in amortized
contexts~\citep{Cremer2018inf}.
It is therefore necessary to test its empirical performance to assert that gains are possible with inexact proposals.
To this end, we 
investigate AMCI's performance on two illustrative examples.

Our primary baseline is the SNIS approach implicitly used by most existing inference amortization methods, namely the
SNIS estimator with proposal $q_2(x;y)$. Though this effectively represents the previous  
state-of-the-art in amortized expectation calculation, it turns out to be a very weak baseline.  We, therefore, introduce another simple
approach one could hypothetically consider using: training separate proposals as per AMCI, but then using
this to form a mixture distribution proposal for an SNIS estimator.  For example, in the scenario where 
$f(x;\theta) \ge 0 \, \forall x,\theta$ (such that we only need to learn two proposals), we can use
\begin{align}
q_{m}(x;y,\theta) = \frac{1}{2}	q_1(x;y,\theta)+\frac{1}{2}	q_2(x;y)
\end{align}
as an SNIS proposal that takes into account the needs of both $E_1$ and $E_2$.  We refer to this
method as the \emph{mixture} SNIS estimator and emphasize that it represents a novel 
amortization approach in its own right.

We also compare AMCI to the theoretically optimal SNIS estimator, i.e. the error bound given
by~\eqref{eq:achievable}.  As we will show, AMCI is often able to empirically outperform this bound, thereby
giving better performance than \emph{any} approach based on SNIS, whether that approach is amortized or not.  This
is an important result and, it particular, it highlights that the potential significance of the AMCI estimator extends beyond the amortized setting we consider here.

We further consider using SNIS with proposal $q_1(x;y,\theta)$.  However, this transpires to perform
extremely poorly throughout (far worse than $q_2(x;y)$) and so we omit its results
from the main paper, giving them in Appendix~\ref{sec:snisq1}.

In all experiments, we use the same number of sample from each proposal to form the estimate (i.e. $N=M=K$).

An implementation for AMCI and our experiments is available at  { \href{http://github.com/talesa/amci}{http://github.com/talesa/amci}}.

\subsection{Tail Integral Calculation}

We start with the conceptually simple problem of calculating tail integrals for Gaussian distributions,
namely
\begin{align}
\label{eq:gaussian-model}
p(x) &= \mathcal{N}(x; 0, \Sigma_1) &
p&(y|x) = \mathcal{N}(y; x, \Sigma_2) \\
f(x; \theta) &= \prod\nolimits_{i=1}^{D} \mathds{1}_{x_i>\theta_i} 
& p&(\theta) = \textsc{Uniform}(\theta;[0,u_D]^D)
\nonumber
\end{align}
where $D$ is the dimensionality, we set $\Sigma_2 = I$, and $\Sigma_1$ is 
a fixed covariance matrix (for details see Appendix~\ref{sec:exp-details}).

This problem was chosen because it permits easy calculation of the ground truth expectations by exploiting analytic simplifications, while remaining numerically challenging for values of $\theta$ far away from the mean when we do not use these simplifications.
We performed one and five-dimensional variants of the experiment.

We use normalizing flows~\citep{Rezende2015} to construct our proposals, providing a flexible and powerful means of representing the target distributions. 
Details are given in Appendix~\ref{sec:exp-details}.
Training was done by using importance sampling to generate the values of $\theta$ and $x$ as per~\eqref{eq:train-prop}
with $q^\prime(\theta, x) = p(\theta) \cdot \textsc{HalfNormal}(x; \theta, \text{diag}(\Sigma_2))$.

To evaluate AMCI and our baselines we use
the relative mean squared error (ReMSE) $\delta(y,\theta) =\E \big[ \hat{\delta}(y,\theta) \big]$, 
where
\begin{align}
\label{eq:rel-error}
\hat{\delta}(y,\theta) = \frac{\left( \mu(y,\theta)-\hat{\mu}(y,\theta) \right)^2}{\mu(y,\theta)^2}
\end{align}
and $\hat{\mu}(y,\theta)$ is our estimate for $\mu(y,\theta)$.
We then consider summary statistics across different $\{y,\theta\}$, such as its median when 
$y,\theta \!\sim \! p(y)p(\theta)$.\footnote{Variability in $\delta(y,\theta)$ between different instances of $\{y,\theta\}$ is considered in Figures~\ref{fig:tails-over-datapoints} and~\ref{fig:cancer-over-datapoints} in Appendix~\ref{sec:snisq1}.}
In calculating this, 
$\delta(y,\theta)$ was separately estimated for
each value of $y$ and $\theta$ using $100$ samples of $\hat{\delta}(y,\theta)$ (i.e. $100$
realizations of the estimator).

As shown in Figure~\ref{fig:tails}, AMCI outperformed SNIS in both the one- and five-dimensional cases.
For the one-dimensional example, AMCI significantly outperformed all of 
SNIS $q_2$, SNIS $q_m$, and the theoretically optimal SNIS estimator.
SNIS $q_2$, the approach implicitly taken by existing inference amortization
methods, typically failed to place even a single sample in the tail of the distribution, even for large $N$.
Interestingly, SNIS $q_m$ closely matched the theoretical SNIS bound, 
suggesting that this amortized proposal is very close to the theoretically
optimal one.  However, this still
constituted significantly worse performance than AMCI---taking about $10^3$ 
more samples to achieve the same relative error---demonstrating the ability
of AMCI to outperform the best possible SNIS estimator.

For the five-dimensional example, AMCI again significantly outperformed 
our main baseline SNIS $q_2$.
Though it still also outperformed SNIS $q_m$, its advantage was less than in one-dimensional case, and it did not outperform the SNIS theoretical bound.
SNIS $q_m$ itself did not match the bound as closely as in the one-dimensional example either, suggesting that the proposals learned were worse than in the one-dimensional case.
Further comparisons based on using the mean squared error (instead of ReMSE) are given in Appendix~\ref{sec:snisq1}
and show qualitatively similar behavior.

\subsection{Planning Cancer Treatment}

\begin{figure}[t]
	\centering
	\begin{minipage}[b]{\columnwidth}
		\includegraphics[width=0.95\columnwidth]{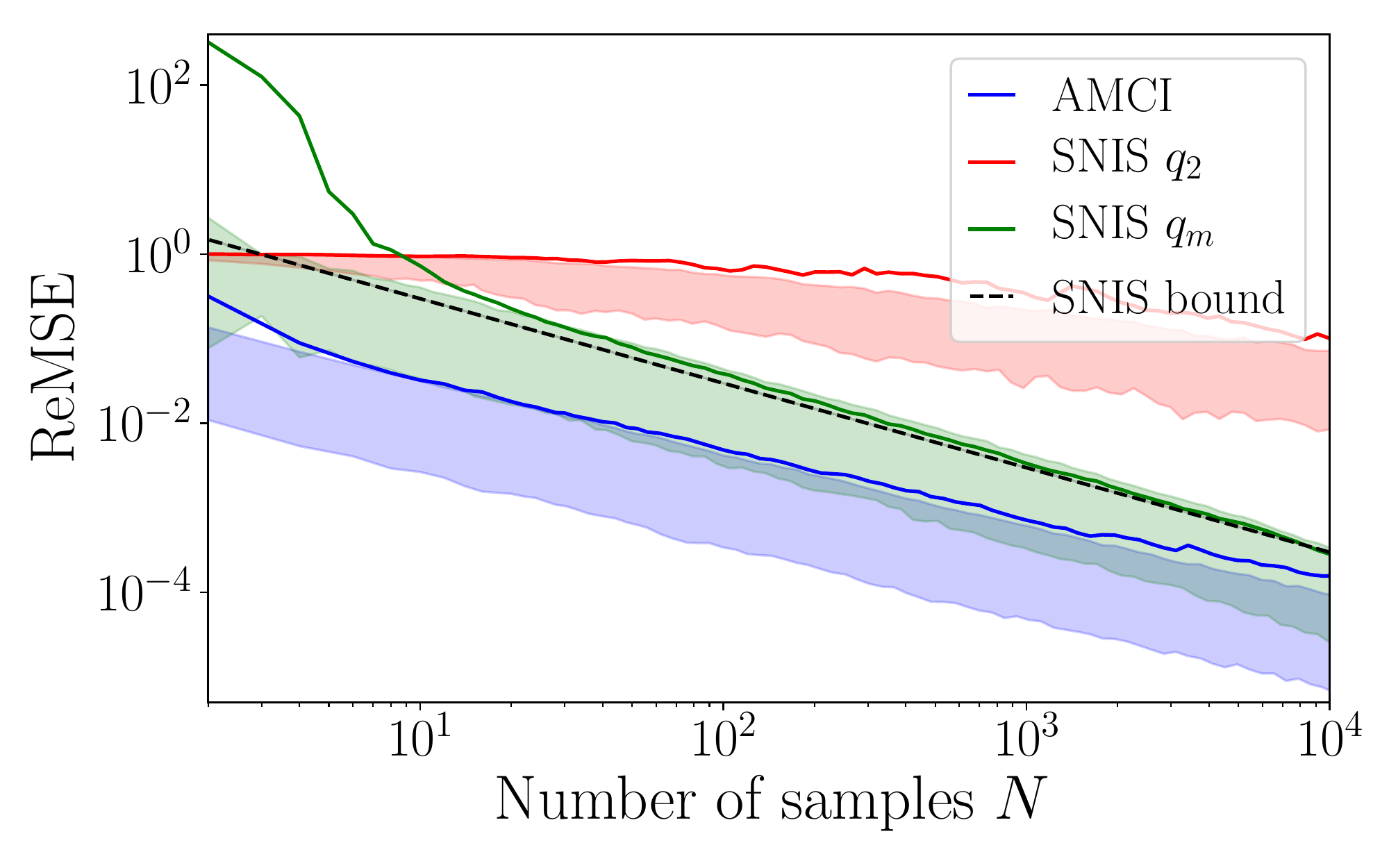}
	\end{minipage}
	\vspace{-22pt}
	\caption{
		Relative mean squared errors for the cancer example.
		Conventions as per Figure~\ref{fig:tails}.
		It is worth noting that it took about $10^4$ more samples for the SNIS $q_2$ estimator to achieve the same level of accuracy as the AMCI estimator. 
	}
	\vspace{-17pt}
	\label{fig:cancer-relative-error-overall}
\end{figure}

To demonstrate how AMCI might be used in a more real-world scenario, we now consider an illustrative example relating
to cancer diagnostic decisions.
Imagine that an oncologist is trying to decide whether to administer a treatment to a cancer patient.
Because the treatment is highly invasive, they only want to administer it if there is a realistic chance of it being successful, i.e. that the tumor shrinks sufficiently to allow a future operation to be carried out.
However, they are only able to make noisy observations about the current size of the tumor, and there are various unknown parameters pertaining to its growth, such as the patients predisposition to the treatment. 
To aid in the oncologists decision, the clinic provides a simulator of tumor evolution, a model of the latent factors required for this simulator, and a loss function for administering the treatment given the final tumor size.
We wish to construct an amortization of this simulator, so that we can quickly and directly predict the expected loss function for administering the treatment from a pair of noisy observations of the tumor size taken at separate points in time. 
A detailed description of the model and proposal setup is in the Appendix~\ref{sec:cancer-model}.

To evaluate the learned proposals we followed the same procedure as for the tail integral example.
Results are presented in Figure~\ref{fig:cancer-relative-error-overall}.
AMCI again significantly outperformed the literature baseline of SNIS $q_2$---it took about $N\!=\!10^4$ samples for SNIS $q_2$ to achieve the level of relative error of AMCI for $N\!=\!2$.
AMCI further maintained an advantage over SNIS $q_m$, which itself
again closely matched the optimal SNIS estimator.
Further comparisons are given in Appendix~\ref{sec:snisq1} and show qualitatively similar behavior.

\section{Discussion}
\label{sec:discussion}

In all experiments AMCI performed better than SNIS with either $q_2$ or $q_m$ for its proposal.
Moreover, it is clear that AMCI is indeed able to break the theoretical bound on the achievable performance
of SNIS estimators: in some cases AMCI is outperforming the best achievable error by any SNIS estimator, regardless of the proposal the latter uses.
Interestingly, the mixture SNIS estimator we also introduce proved to be a strong baseline as it closely 
matched the theoretical baseline in both experiments.
However, such an effective mixture proposal is only possible thanks learning the multiple inference artifacts
we suggest as part of the AMCI framework, while its performance was still generally inferior to AMCI itself.

We now consider the question of when we expect AMCI to work particularly well compared to SNIS, and the scenarios where it is less beneficial, or potentially even harmful.
We first note that scaling with increasing dimensionality is a challenge for both because the importance
sampling upon which they rely suffers from the curse of dimensionality.
However, the scaling of AMCI should be no worse than existing amortization approaches as each of the amortized proposals is trained in isolation and corresponds to a conventional inference amortization.

We can gain more insights into the relative performance of the two approaches in different settings using an informal asymptotic analysis in the limit of a large number of samples.
Assuming $f(x;\theta)\ge0 \, \forall x, \theta$ for simplicity,\footnote{The results trivially generalize to general $f(x)$ with suitable adjustment of the definition of $\sigma_1$.}
then both AMCI and SNIS can be expressed in the form of~\eqref{eq:amci-mc-estimator}, where for SNIS we set $q_1(x;y,\theta)\!=\!q_2(x;y)$, $N\!=\!M$, and share samples between the estimators.
Separately applying the central limit theorem to $\hat{E}_1$ and $\hat{E}_2$ yields
\begin{align}
  \hat{\mu}(y,\theta) = \frac{\hat{E}_1}{\hat{E}_2}\to&~ \frac{E_1 + \sigma_1 \xi_1}{E_2 + \sigma_2 \xi_2}, \quad 
  \text{as} \quad N,M\to\infty
\end{align}
where $\xi_1, \xi_2 \sim \mathcal{N}(0,1)$ and
\begin{align}
\sigma_1 &:= \frac{1}{N}\text{Var}_{q_1(x;y,\theta)}\left[\frac{f(x;\theta)p(x,y)}{q_1(x;y,\theta)}\right], \displaybreak[0] \\
\sigma_2 &:= \frac{1}{M}\text{Var}_{q_2(x;y)}\left[\frac{p(x,y)}{q_2(x;y)}\right].
\end{align}
Asymptotically, the mean squared error of $\hat{\mu}(y,\theta)$ is dominated by its variance. 
Thus, by taking a first order Taylor expansion 
of $\text{Var}[\hat{\mu}(y,\theta)]$ about $1/E_2$, we get, for large $M$,
\begin{align}
	&\hspace{-7pt}\E \left[\left(\hat{\mu}(y,\theta)-\mu(y,\theta)\right)^2\right] \nonumber \\
	&\hspace{-7pt}\approx \,\, \frac{1}{E_2^2} \left( \sigma_1^2 + \sigma_2^2 \mu(y,\theta)^2 - 2 \mu(y,\theta) \sigma_1 \sigma_2  \text{Corr}[\xi_1, \xi_2] \right) \nonumber \displaybreak[0]\\
	&\hspace{-7pt}=\frac{\sigma_2^2}{E_2^2} \left( (\kappa - \text{Corr}[\xi_1, \xi_2])^2 + 1 - \text{Corr}[\xi_1, \xi_2]^2 \right)
\end{align}
where the approximation from the Taylor expansion becomes exact in the limit $M \to \infty$ and $\kappa:=\sigma_1/(\mu(y,\theta)\sigma_2)$ is a measure of the relative accuracy of the two estimators.
See \eqref{eq:amci-mu-variance} in \hyperref[sec:opt-vals]{Appendix~\ref*{sec:opt-vals}} for a more verbose derivation.

For a given value of $\sigma_2$, the value of $\kappa$ for SNIS is completed dictated by the problem: in general, the larger the mismatch between $f(x;\theta)p(x,y)$ and $p(x,y)$, the larger $\kappa$ will be.
This yields the expected result that the errors for SNIS become large in this setting.
For AMCI, we can control $\kappa$ through ensuring a good proposal for both $\hat{E}_1$ and $\hat{E}_2$, and, if desired, by adjusting $M$ and $N$ (relative to a fixed budget $M+N$).
Consequently, we can achieve better errors than SNIS by driving $\kappa$ down.

On the other hand, as $f(x;\theta)p(x,y)$ and $p(x,y)$ become increasingly well matched, then $\kappa\to1$ and we find that AMCI has little to gain over SNIS.
In fact, we see that AMCI can potentially be worse than SNIS in this setting:
when $f(x;\theta)p(x,y)$ and $p(x,y)$ are closely matched, we also have $\text{Corr}[\xi_1, \xi_2]^2\approx 1$ for SNIS, such that we observe a canceling effect, potentially leading to very low errors.
Achieving $\text{Corr}[\xi_1, \xi_2]^2\approx 1$ can be more difficult for AMCI, potentially giving rise to a higher error.
However, it could be possible to mitigate this by correlating the estimates, e.g. through common random numbers.

\begin{figure}[t!]
	\centering
	\adjincludegraphics[width=0.96\columnwidth,trim={{.01\width} 0 {.35\width} 0},clip]{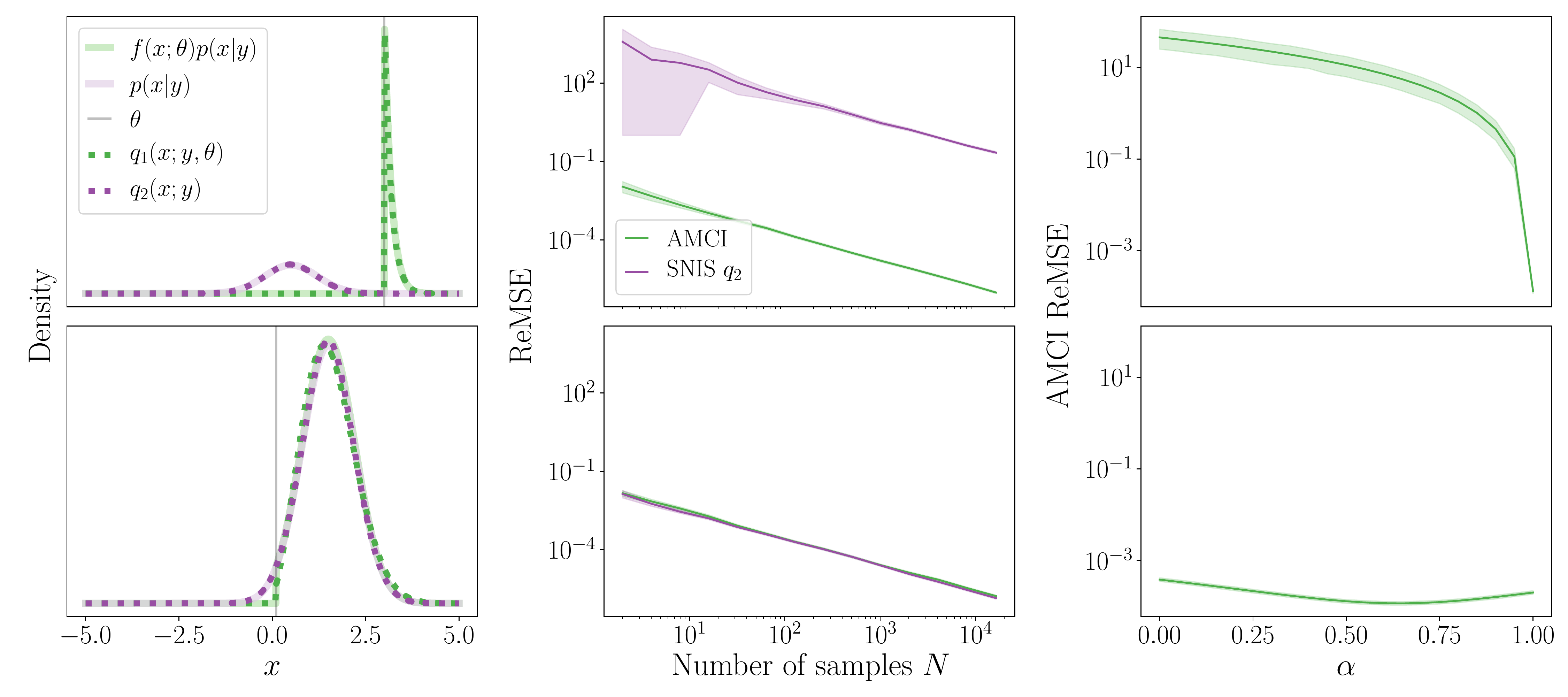}
	\vspace{-10pt}
	\caption{
		Results for the one-dimensional tail integral model in a setting with large mismatch [top] and low mismatch [bottom], with $(y, \theta)$, respectively $(1, 3)$ and $(3, 0.1)$.
		The left column illustrates the shape of the proposal $q_1$ and the achievable quality of fit to $f(x; \theta)p(x|y)$, we see that AMCI is able to learn very accurate proposals in both cases.
		The right column compares the performance of the AMCI and the SNIS estimators where we see that the gain for AMCI is much larger when the mismatch is large.
		Uncertainty bands in column two are estimated over a 1000 runs and are almost imperceptibly small.
	}
  \vspace{-12pt}
	\label{fig:tail-1d-figure}
\end{figure}

To assess if this theory manifests in practice, we revisit our tail integral example, comparing large and small mismatch scenarios. 
The results, shown in Figure~\ref{fig:tail-1d-figure}, agree with these theoretical findings.  
In Appendix~\ref{sec:reusing-samples} we further showing that the reusing of samples for both $\hat{E}_1$ and $\hat{E}_2$ in AMCI can be beneficial when the targets are well matched.

More generally, as Theorem 1 tells us that the AMCI estimator can achieve an arbitrarily low error for any given target function, while SNIS cannot, we know that its potential gains are larger the more accurate we are able to make our proposals.
As such, as advances elsewhere in the field allow us to produce increasingly effective amortized proposals, e.g. through advanced normalizing flow approaches \citep{Grathwohl2018, Kingma2018glow}, the larger the potential gains are from using AMCI.

\vfill
\clearpage

\section*{Acknowledgments}
We would like to thank Yee Whye Teh for providing helpful discussions at the early stages of the project. 
AG is supported by the UK EPSRC CDT in Autonomous Intelligent Machines and Systems. 
FW is supported by DARPA D3M, under Cooperative Agreement FA8750-17-2-0093, Intel under its LBNL NERSC Big Data Center, and an NSERC Discovery grant.
TR is supported by the European Research Council under the European Union’s Seventh Framework Programme (FP7/2007–2013) / ERC grant agreement no. 617071.
His research leading to these results also received funding from EPSRC under grant EP/P026753/1.

\bibliography{bibliography/standardstrings,bibliography/longstrings,bibliography/bibliography,project-specific}

\begin{thebibliography}{35}
\providecommand{\natexlab}[1]{#1}
\providecommand{\url}[1]{\texttt{#1}}
\expandafter\ifx\csname urlstyle\endcsname\relax
  \providecommand{\doi}[1]{doi: #1}\else
  \providecommand{\doi}{doi: \begingroup \urlstyle{rm}\Url}\fi

\bibitem[Bugallo et~al.(2017)Bugallo, Elvira, Martino, Luengo, Miguez, and
  Djuric]{Bugallo2017}
Bugallo, M.~F., Elvira, V., Martino, L., Luengo, D., Miguez, J., and Djuric,
  P.~M.
\newblock Adaptive importance sampling: the past, the present, and the future.
\newblock \emph{IEEE Signal Processing Magazine}, 34\penalty0 (4):\penalty0
  60--79, 2017.

\bibitem[Chen \& Shao(1997)Chen and Shao]{Chen1997}
Chen, M.-H. and Shao, Q.-M.
\newblock On {M}onte {C}arlo methods for estimating ratios of normalizing
  constants.
\newblock \emph{The Annals of Statistics}, 25\penalty0 (4):\penalty0
  1563--1594, 08 1997.

\bibitem[Cremer et~al.(2018)Cremer, Li, and Duvenaud]{Cremer2018inf}
Cremer, C., Li, X., and Duvenaud, D.
\newblock Inference suboptimality in variational autoencoders.
\newblock \emph{Proceedings of the International Conference on Machine Learning
  (ICML)}, 2018.

\bibitem[Enderling \& Chaplain(2014)Enderling and Chaplain]{Enderling2014}
Enderling, H. and Chaplain, M.~A.
\newblock Mathematical modeling of tumor growth and treatment.
\newblock \emph{Current pharmaceutical design}, 20--30:\penalty0 4934--40,
  2014.

\bibitem[Evans \& Swartz(1995)Evans and Swartz]{Evans1995}
Evans, M. and Swartz, T.
\newblock Methods for approximating integrals in statistics with special
  emphasis on {B}ayesian integration problems.
\newblock \emph{Statistical science}, pp.\  254--272, 1995.

\bibitem[Gelman \& Meng(1998)Gelman and Meng]{Gelman1998}
Gelman, A. and Meng, X.-L.
\newblock Simulating normalizing constants: from importance sampling to bridge
  sampling to path sampling.
\newblock \emph{Statistical Science}, 13\penalty0 (2):\penalty0 163--185, 05
  1998.

\bibitem[Grathwohl et~al.(2019)Grathwohl, Chen, Bettencourt, Sutskever, and
  Duvenaud]{Grathwohl2018}
Grathwohl, W., Chen, R. T.~Q., Bettencourt, J., Sutskever, I., and Duvenaud, D.
\newblock {FFJORD:} free-form continuous dynamics for scalable reversible
  generative models.
\newblock \emph{International Conference on Learning Representations (ICLR)},
  2019.

\bibitem[Hahnfeldt et~al.(1999)Hahnfeldt, Panigrahy, Folkman, and
  Hlatky]{Hahnfeldt1999}
Hahnfeldt, P., Panigrahy, D., Folkman, J., and Hlatky, L.
\newblock Tumor development under angiogenic signaling.
\newblock \emph{Cancer Research}, 59\penalty0 (19):\penalty0 4770--4775, 1999.

\bibitem[Hesterberg(1988)]{Hesterberg1988}
Hesterberg, T.~C.
\newblock \emph{Advances in importance sampling}.
\newblock PhD thesis, Stanford University, 1988.

\bibitem[Hoffman et~al.(2013)Hoffman, Blei, Wang, and Paisley]{Hoffman2013}
Hoffman, M.~D., Blei, D.~M., Wang, C., and Paisley, J.
\newblock Stochastic variational inference.
\newblock \emph{Journal of Machine Learning Research (JMLR)}, 2013.

\bibitem[Kingma \& Ba(2015)Kingma and Ba]{Kingma2015}
Kingma, D.~P. and Ba, J.
\newblock Adam: A method for stochastic optimization.
\newblock \emph{International Conference on Learning Representations (ICLR)},
  2015.

\bibitem[Kingma \& Dhariwal(2018)Kingma and Dhariwal]{Kingma2018glow}
Kingma, D.~P. and Dhariwal, P.
\newblock Glow: Generative flow with invertible 1x1 convolutions.
\newblock \emph{Advances in Neural Information Processing Systems (NIPS)},
  2018.

\bibitem[Kingma \& Welling(2014)Kingma and Welling]{Kingma2014}
Kingma, D.~P. and Welling, M.
\newblock Auto-encoding variational {B}ayes.
\newblock \emph{International Conference on Learning Representations (ICLR)},
  2014.

\bibitem[Lacoste-Julien et~al.(2011)Lacoste-Julien, Husz{\'a}r, and
  Ghahramani]{Lacoste-Julien2011}
Lacoste-Julien, S., Husz{\'a}r, F., and Ghahramani, Z.
\newblock Approximate inference for the loss-calibrated {B}ayesian.
\newblock \emph{Proceedings of the International Conference on Artificial
  Intelligence and Statistics (AISTATS)}, 2011.

\bibitem[Lamberti et~al.(2018)Lamberti, Petetin, Septier, and
  Desbouvries]{Lamberti2018}
Lamberti, R., Petetin, Y., Septier, F., and Desbouvries, F.
\newblock A double proposal normalized importance sampling estimator.
\newblock \emph{2018 IEEE Statistical Signal Processing Workshop (SSP)}, pp.\
  238--242, 2018.

\bibitem[Le et~al.(2017)Le, Baydin, and Wood]{Le2016}
Le, T.~A., Baydin, A.~G., and Wood, F.
\newblock Inference compilation and universal probabilistic programming.
\newblock \emph{Proceedings of the International Conference on Artificial
  Intelligence and Statistics (AISTATS)}, 2017.

\bibitem[Le et~al.(2018{\natexlab{a}})Le, Igl, Jin, Rainforth, and
  Wood]{Le2017}
Le, T.~A., Igl, M., Jin, T., Rainforth, T., and Wood, F.
\newblock Auto-encoding sequential {M}onte {C}arlo.
\newblock \emph{International Conference on Learning Representations (ICLR)},
  2018{\natexlab{a}}.

\bibitem[Le et~al.(2018{\natexlab{b}})Le, Kosiorek, Siddharth, Teh, and
  Wood]{Le2018}
Le, T.~A., Kosiorek, A.~R., Siddharth, N., Teh, Y.~W., and Wood, F.
\newblock Revisiting reweighted wake-sleep.
\newblock \emph{arXiv:1805.10469}, 2018{\natexlab{b}}.

\bibitem[Meng \& Wong(1996)Meng and Wong]{Meng1996}
Meng, X.-L. and Wong, W.~H.
\newblock Simulating ratios of normalizing constants via a simple identity: A
  theoretical exploration.
\newblock \emph{Statistica Sinica}, 6:\penalty0 831--860, 1996.

\bibitem[Oh \& Berger(1992)Oh and Berger]{Oh1992}
Oh, M.-S. and Berger, J.~O.
\newblock Adaptive importance sampling in {M}onte {C}arlo integration.
\newblock \emph{Journal of Statistical Computation and Simulation}, 41\penalty0
  (3-4):\penalty0 143--168, 1992.

\bibitem[Owen(2013)]{Owen2013}
Owen, A.~B.
\newblock \emph{{M}onte {C}arlo theory, methods and examples}.
\newblock 2013.

\bibitem[Paige \& Wood(2016)Paige and Wood]{Paige2016}
Paige, B. and Wood, F.
\newblock Inference networks for sequential {M}onte {C}arlo in graphical
  models.
\newblock \emph{Proceedings of the International Conference on Machine Learning
  (ICML)}, 2016.

\bibitem[Papamakarios et~al.(2017)Papamakarios, Pavlakou, and
  Murray]{Papamakarios2017}
Papamakarios, G., Pavlakou, T., and Murray, I.
\newblock Masked autoregressive flow for density estimation.
\newblock \emph{Advances in Neural Information Processing Systems (NIPS)},
  2017.

\bibitem[Rainforth(2017)]{Rainforth2017}
Rainforth, T.
\newblock \emph{Automating inference, learning, and design using probabilistic
  programming}.
\newblock PhD thesis, 2017.

\bibitem[Rainforth et~al.(2018{\natexlab{a}})Rainforth, Cornish, Yang,
  Warrington, and Wood]{Rainforth2018nmc}
Rainforth, T., Cornish, R., Yang, H., Warrington, A., and Wood, F.
\newblock {On Nesting Monte Carlo Estimators}.
\newblock \emph{Proceedings of the International Conference on Machine Learning
  (ICML)}, 2018{\natexlab{a}}.

\bibitem[Rainforth et~al.(2018{\natexlab{b}})Rainforth, Zhou, Lu, Teh, Wood,
  Yang, and van~de Meent]{Rainforth2018it}
Rainforth, T., Zhou, Y., Lu, X., Teh, Y.~W., Wood, F., Yang, H., and van~de
  Meent, J.-W.
\newblock Inference trees: Adaptive inference with exploration.
\newblock \emph{arXiv preprint arXiv:1806.09550}, 2018{\natexlab{b}}.

\bibitem[Rezende \& Mohamed(2015)Rezende and Mohamed]{Rezende2015}
Rezende, D. and Mohamed, S.
\newblock Variational inference with normalizing flows.
\newblock \emph{Proceedings of the International Conference on Machine Learning
  (ICML)}, 2015.

\bibitem[Rezende et~al.(2014)Rezende, Mohamed, and Wierstra]{Rezende2014}
Rezende, D.~J., Mohamed, S., and Wierstra, D.
\newblock Stochastic backpropagation and approximate inference in deep
  generative models.
\newblock \emph{Proceedings of the International Conference on Machine Learning
  (ICML)}, 2014.

\bibitem[Ritchie et~al.(2016)Ritchie, Horsfall, and Goodman]{Ritchie2016}
Ritchie, D., Horsfall, P., and Goodman, N.~D.
\newblock Deep amortized inference for probabilistic programs.
\newblock \emph{arXiv:1610.05735}, 2016.

\bibitem[Robert(2007)]{Robert2007}
Robert, C.
\newblock \emph{The {B}ayesian choice: from decision-theoretic foundations to
  computational implementation}.
\newblock Springer Science \& Business Media, 2007.

\bibitem[Robert \& Casella(2013)Robert and Casella]{Robert2013}
Robert, C. and Casella, G.
\newblock \emph{{M}onte {C}arlo statistical methods}.
\newblock Springer Science \& Business Media, 2013.

\bibitem[Stuhlm{\"u}ller et~al.(2013)Stuhlm{\"u}ller, Taylor, and
  Goodman]{Stuhlmuller2013}
Stuhlm{\"u}ller, A., Taylor, J., and Goodman, N.
\newblock Learning stochastic inverses.
\newblock \emph{Advances in Neural Information Processing Systems (NIPS)},
  2013.

\bibitem[Veach \& Guibas(1995)Veach and Guibas]{Veach1995}
Veach, E. and Guibas, L.~J.
\newblock Optimally combining sampling techniques for {M}onte {C}arlo
  rendering.
\newblock \emph{Proceedings of the 22nd annual conference on Computer graphics
  and interactive techniques}, pp.\  419--428, 1995.

\bibitem[Webb et~al.(2018)Webb, Goli{\'n}ski, Zinkov, Siddharth, Rainforth,
  Teh, and Wood]{Webb2018}
Webb, S., Goli{\'n}ski, A., Zinkov, R., Siddharth, N., Rainforth, T., Teh,
  Y.~W., and Wood, F.
\newblock Faithful inversion of generative models for effective amortized
  inference.
\newblock \emph{Advances in Neural Information Processing Systems (NIPS)},
  2018.

\bibitem[Wolpert(1991)]{Wolpert1991}
Wolpert, R.~L.
\newblock {M}onte {C}arlo integration in {B}ayesian statistical analysis.
\newblock \emph{Contemporary Mathematics}, 115:\penalty0 101--116, 1991.

\end{thebibliography}
\bibliographystyle{icml2019}

\vfill\clearpage
\appendix
\onecolumn

	\thispagestyle{empty} 
	\rule{\textwidth}{1pt}
	\vspace{-6pt}
	\begin{center}
		\textbf{ \Large Appendices for Amortized Monte Carlo Integration}
	\end{center}\vspace{-6pt}
	\rule{\textwidth}{1pt}
	\begin{minipage}{\textwidth}
		\centering
		\vspace{17pt}
		\textbf{Adam Goli{\'n}ski* \quad Frank Wood \quad Tom Rainforth*}
		\vspace{-2pt}
	\end{minipage}

\vspace{10pt}

\section{Additional Experimental Results}
\label{sec:snisq1}

\begin{figure}[H]
	\centering
	\begin{minipage}[b]{\linewidth}
		\includegraphics[width=\textwidth]{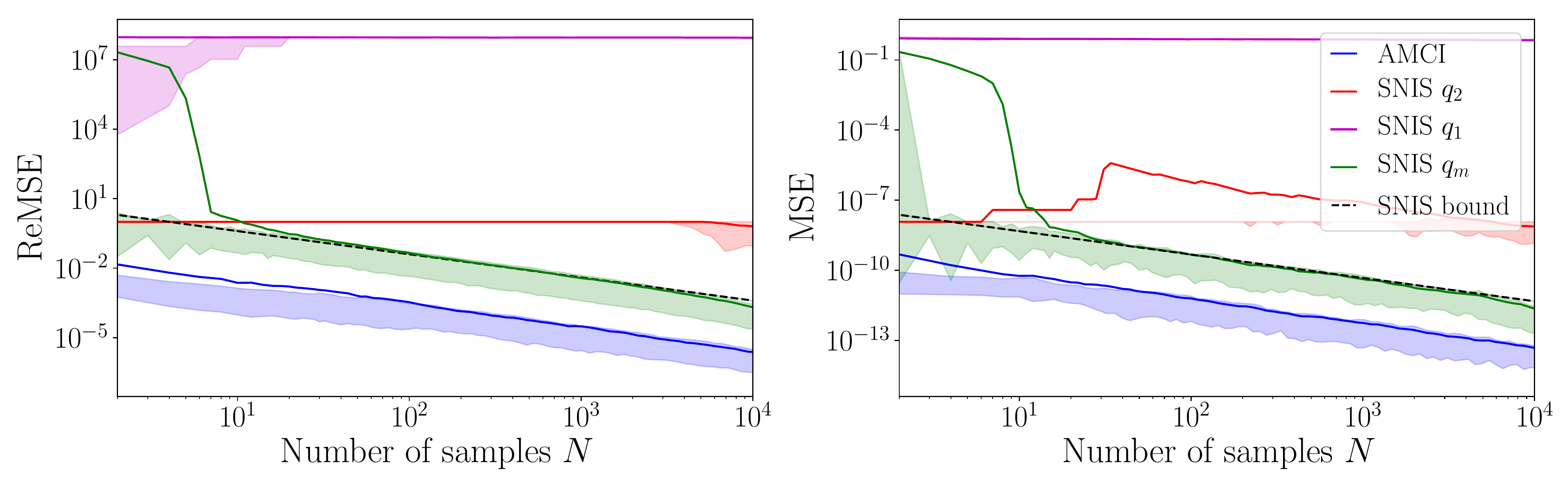}
	\end{minipage}
	\vspace{-15pt}
	\caption{
		Additional results for one-dimensional tail integral example as per Figure~\ref{fig:one-dim}.
	[left] Relative mean squared errors (as per~\eqref{eq:rel-error}). 
	[right] Mean squared error $\E [(\mu(y,\theta)-\hat{\mu}(y,\theta)^2]$.
	Conventions as per Figure~\ref{fig:tails}.
	The results for SNIS $q_1$ indicate that it severely underestimates $E_2$ leading to very large errors, especially when the mismatch between $p(x|y)$ and $f(x; \theta)$ is as significant as in the tail integral case.
	}
	\vspace{-15pt}
	\label{fig:tail1d_q1}
\end{figure}

\begin{figure}[H]
	\centering
	\begin{minipage}[b]{\linewidth}
		\includegraphics[width=\textwidth]{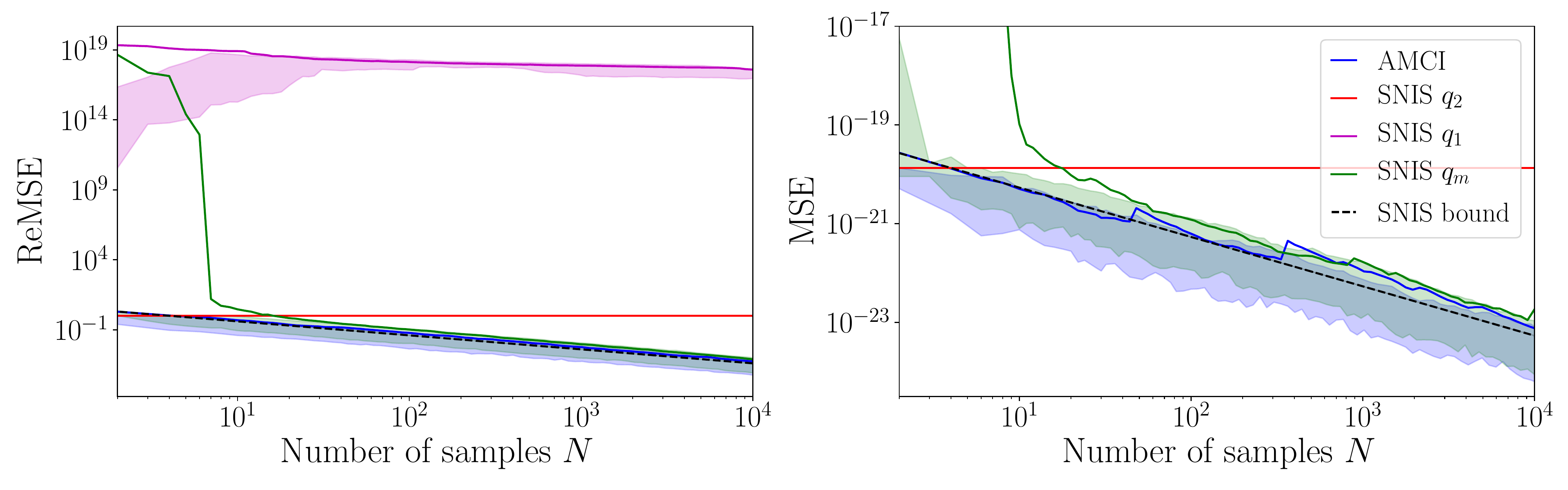}
	\end{minipage}
	\vspace{-15pt}
	\caption{
		Additional results for five-dimensional tail integral example as per Figure~\ref{fig:five-dim}.
	[left] Relative mean squared errors (as per~\eqref{eq:rel-error}). 
	[right] Mean squared error $\E [(\mu(y,\theta)-\hat{\mu}(y,\theta)^2]$.
	Conventions as per Figure~\ref{fig:tails}.
	  The y-axis limits for the MSE have been readjusted to allow clear comparison at higher $N$.
	Note that the SNIS $q_m$ yields MSE of $10^{-1}$ at $N\!=\!2$, while  
	the SNIS $q_1$ MSE is far away from the range of the plot for all $N$, giving a MSE
	of $10^{-0.9}$ at $N\!=\!2$ and $10^{-1.2}$ at $N\!=\!10^4$,
	with a shape very similar to the ReMSE for SNIS $q_1$ as  per the left plot.
	The extremely high errors for SNIS $q_m$ at low values of $N$ arise in the situation when all $N$ samples drawn happen to come from distribution $q_1$.
	We believe that the results presented for $q_m$ underestimate the value of $\delta(y, \theta)$ between around $N=6$ and $N=100$, due to the fact that the estimation process for $\delta(y, \theta)$, though unbiased, can have a very large skew.
	For $N \le 6$ there is a good chance of at least one of the $100$ trials we perform having all $N$ samples originating from distribution $q_1$, such that we generate reasonable estimates for the very high errors this can induce.
	For $N \ge 100$ the chances of this event occurring drop to below $10^{-30}$, such that it does not substantially influence the true error.
	For $6 \le N \le 100$, the chance the event will occur in our $100$ trials is small, but the influence it has on the overall error is still significantly, meaning it is likely we will underestimate the error.
	This effect could be alleviated by Rao-Blackwellizing the choice of the mixture component, but this would induce a stratified sampling estimate, thereby moving beyond the SNIS framework.
	}
	\vspace{-15pt}
	\label{fig:tail5d_q1}
\end{figure}

\begin{figure}[H]
	\centering
	\begin{minipage}[b]{\linewidth}
		\includegraphics[width=\textwidth]{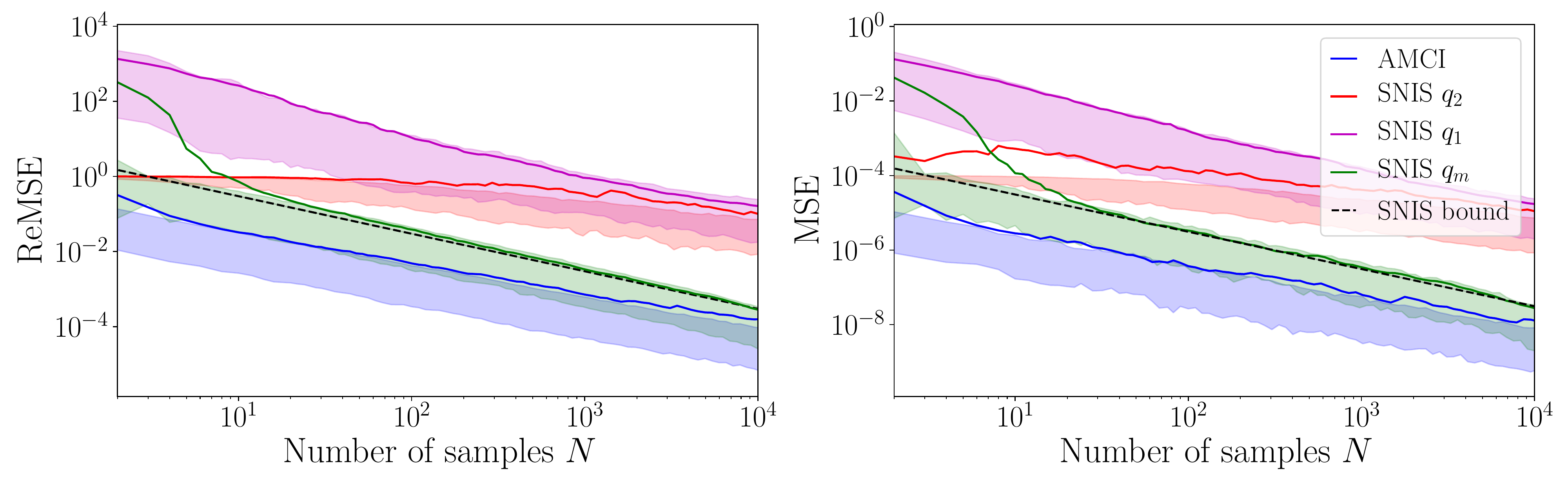}
	\end{minipage}
	\vspace{-15pt}
	\caption{
				Additional results for cancer example as per Figure~\ref{fig:cancer-relative-error-overall}.
				[left] Relative mean squared errors (as per~\eqref{eq:rel-error}). 
				[right] Mean squared error $\E [(\mu(y,\theta)-\hat{\mu}(y,\theta)^2]$.
	Conventions as per Figure~\ref{fig:tails}.
	Here, the SNIS $q_1$ performs much better than in the tail integral example because
	of smaller mismatch between $p(x|y)$ and $f(x; \theta)$, meaning the estimates for $E_2$ are more reasonable.
	Nonetheless, we see that SNIS $q_1$ still performs worse that even SNIS $q_2$.
	}
	\vspace{-15pt}
	\label{fig:cancer-3-conv}
\end{figure}

\begin{figure}[H]
	\centering
	\begin{subfigure}{0.49\columnwidth}
		\centering
		\includegraphics[width=\columnwidth]{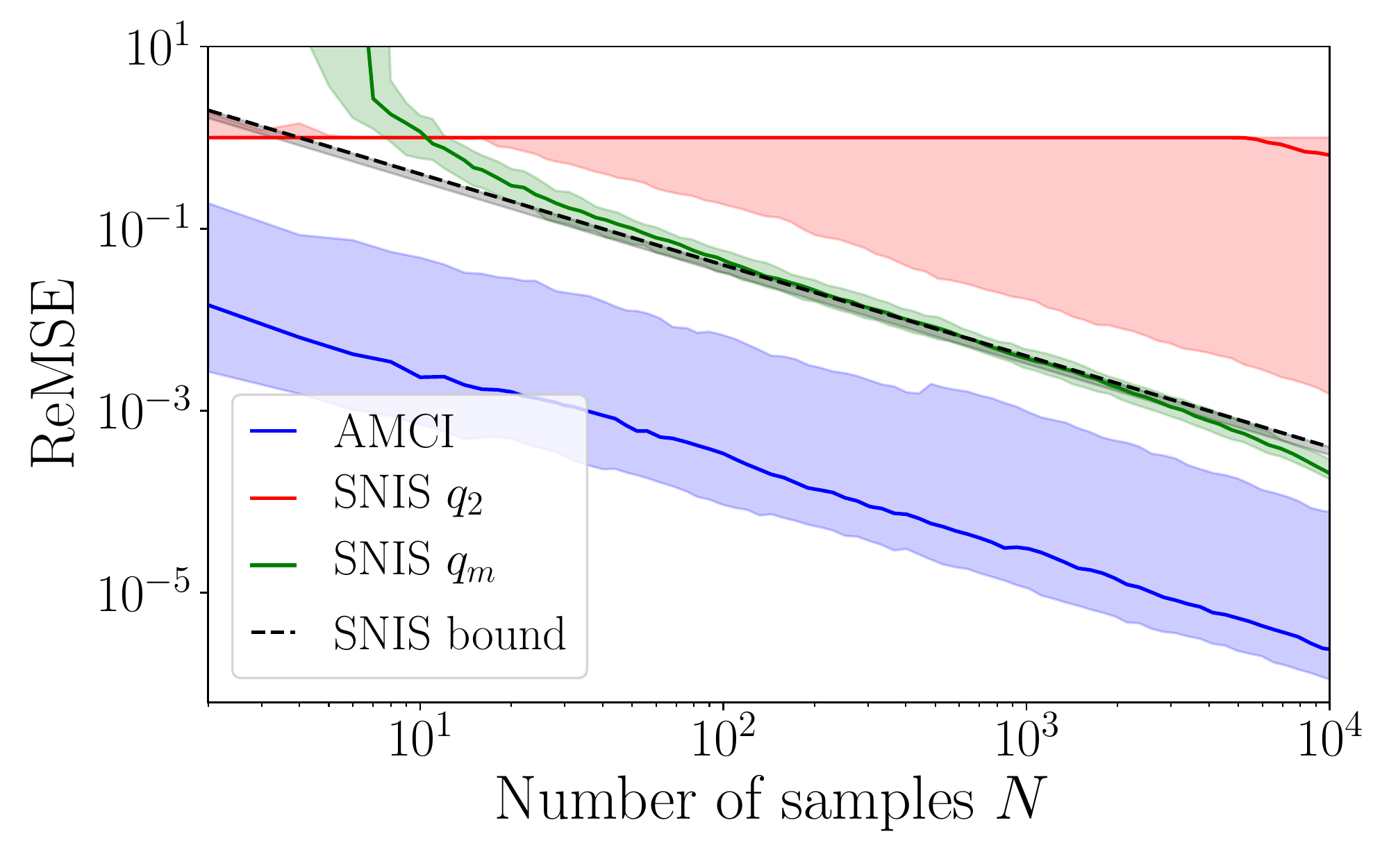}
		\subcaption{One-dimensional}
	\end{subfigure}
	\begin{subfigure}{0.49\columnwidth}
		\centering
		\includegraphics[width=\columnwidth]{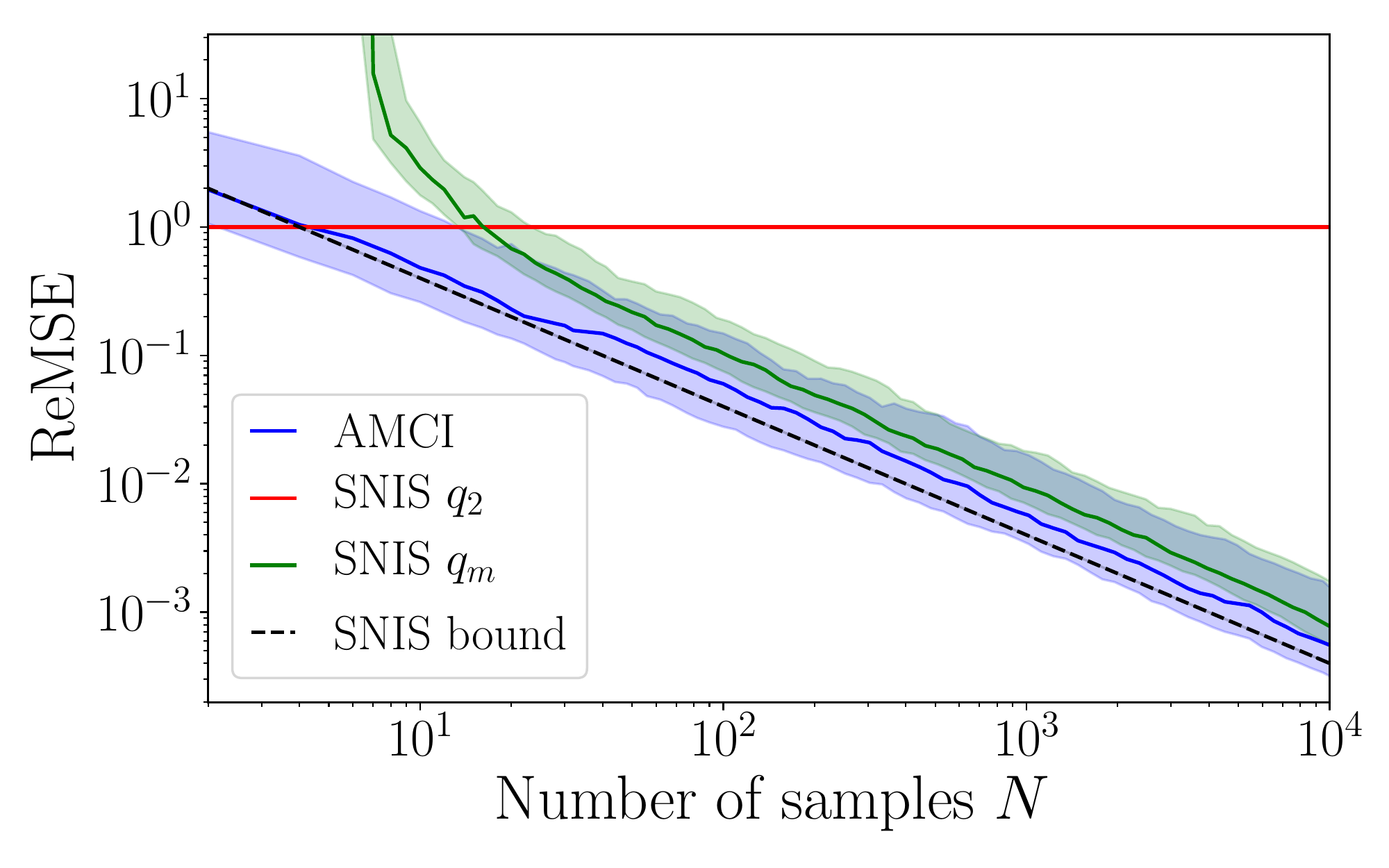}
		\subcaption{Five-dimensional}
	\end{subfigure}
	\caption{
		Investigation of the variability of the results across datapoints $y, \theta$
		for [left] the one-dimensional and [right] the five-dimensional tail integral example.
		Unlike previous figures, 
		the shading shows the estimates of the $25\%$ and $75\%$ quantiles of $\delta(y,\theta)$ estimated using a
		common set of $100$ samples from $y,\theta\sim p(y)p(\theta)$, with the corresponding $\delta(y,\theta)$
		then each separately estimated using $100$ samples of the respective $\hat{\delta}(y, \theta)$.  
		The solid lines for each estimator and the dashed line remain the same as in previous figures -- they indicate the median of $\delta(y,\theta)$.
		Now the dashed line also has a shaded area associated with it reflecting the variability in the SNIS bound across datapoints.
	}
	\vspace{-15pt}
	\label{fig:tails-over-datapoints}
\end{figure}

\begin{figure}[H]
	\centering
	\begin{minipage}{0.49\columnwidth}
		\centering
		\includegraphics[width=\columnwidth]{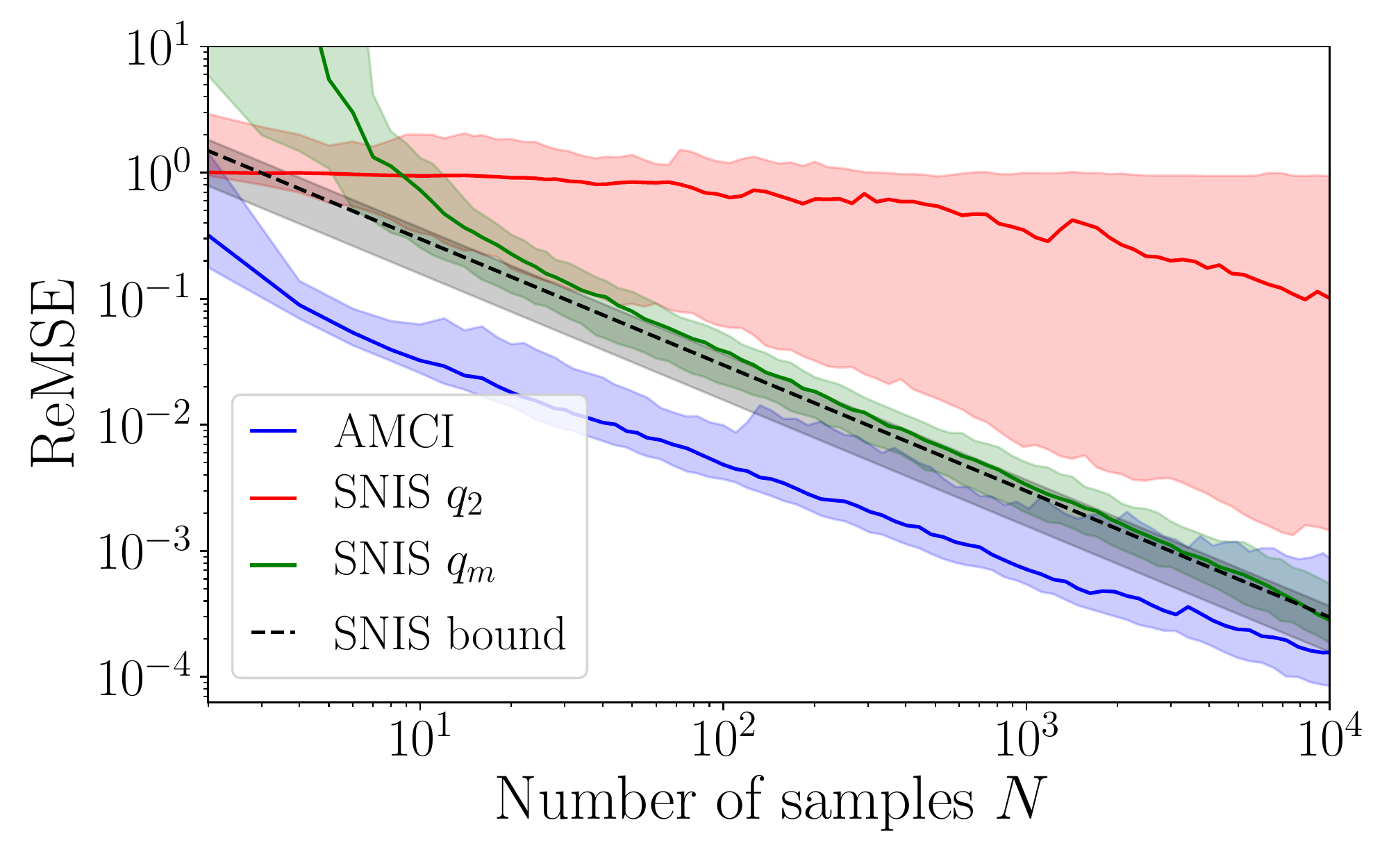}
	\end{minipage}
	\vspace{-15pt}
	\caption{
		Investigation of the variability of the results across datapoints $y, \theta$ for cancer example.
		Conventions as per Figure~\ref{fig:tails-over-datapoints}.
		The fact that the upper quantile of the AMCI error is larger than the upper quantile of the SNIS $q_m$ error
		suggests that there are datapoints for which AMCI yields higher mean squared error than SNIS $q_m$.
		However, AMCI is still always better than the standard baseline, i.e. SNIS $q_2$.
	}
	\vspace{-15pt}
	\label{fig:cancer-over-datapoints}
\end{figure}

\section{Proof of Theorem~\ref{theorem}}
\label{sec:proof}

\amciThe*
\vspace{-10pt}
\begin{proof}
The result follows straightforwardly from considering each estimator in isolation.
Note that the normalization constants for distributions 
$q_1^+, q_1^-, q_2$ are $E_1^+, E_1^-, E_2$, respectively, 
e.g. $\int \! f^+(x^+;\theta)p(x^+,y) \diff x^+ \!=\! E_1^+$.
Therefore, starting with $\hat{E}_2$, we have 
\begin{align}
	\hspace{-4pt}
\hat{E}_2 
&\!=\! \frac{1}{M} \!\!\sum_{m=1}^M \!\frac{p(x_m, y)}{q_2(x_m;y)}
 \!=\! \frac{1}{M} \!\!\sum_{m=1}^M \!\frac{p(x_m, y)}{p(x_m, y)/E_2}
\!=\! E_2
\end{align}
for all possible values of $x_m$. 
Similarly, for $\hat{E}_1^+$
\begin{align}
\hat{E}^+_1
&\!=\! \frac{1}{N} \!\sum_{n=1}^N \frac{p(x_n^+, y)f^+(x_n^+; \theta)}{q_1(x_n^+;y, \theta)} 
\!=\! \frac{1}{N} \!\sum_{n=1}^N \frac{p(x_n^+, y)f^+(x_n^+; \theta)}{p(x_n^+, y)f^+(x_n^+; \theta)/E^+_1}
\!=\! E^+_1
\end{align}
for all possible values of $x_n^+$. 
Analogously, we have $\hat{E}_1^-=E_1^-$ for all possible values of $x_k^-$.
Combining all of the above, the result now follows.
\end{proof}

\section{Experimental details}
\label{sec:exp-details}

\subsection{One-dimensional tail integral}

Let us recall the model from \eqref{eq:gaussian-model},
\begin{align}
p(x) &= \mathcal{N}(x; 0, \Sigma_1) &
p&(y|x) = \mathcal{N}(y; x, \Sigma_2) &
f(x; \theta) &= \prod\nolimits_{i=1}^{D} \mathds{1}_{x_i>\theta_i} 
& p&(\theta) = \textsc{Uniform}(\theta;[0,u_D]^D)
\nonumber
\end{align}
where for the one-dimensional example $D=1$ we used $u_1 = 5$ and $\Sigma_1 = \Sigma_2 = 1$.

For our parameterized proposals $q_1(x;y, \theta)$ and $q_2(x;y)$ we used a normalizing flow consisting of $10$ radial flow layers \citep{Rezende2015} with a standard normal base distribution.
The parameters of each flow were determined by a neural network taking in the values of $y$ and $\theta$ as input, and returning
the parameters defining the flow transformations.
Each network comprised of 3 fully connected layers with 1000 hidden units each layer, with relu activation functions.

Training was done by using importance sampling to generate the values of $\theta$ and $x$ as per~\eqref{eq:train-prop}
with \[
q^\prime(\theta, x) = p(\theta) \cdot \textsc{HalfNormal}(x; \mu=\theta, \sigma=\Sigma_2).
\]
and a learning rate of $10^{-2}$ with the Adam optimizer \citet{Kingma2015}.

The ground truth values of $\mu(y, \theta)$ were determined analytically using $\mu(y, \theta) = \mathbb{E}_{p(x|y)} \!\left[ f(x; \theta) \right] \! = \! 1-\Phi(\theta)$, where $\Phi(\cdot)$ is the standard normal cumulative distribution function.

\subsection{Five-dimensional tail integral}

In the context of the model definition in \eqref{eq:gaussian-model}, 
for the five-dimensional example we used $u_5 = 3$, $\Sigma_2 = I$ and
\[
\Sigma_1 =
\begin{bmatrix}
1.2449 & 0.2068 & 0.1635 & 0.1148 & 0.0604 \\
0.2068 & 1.2087 & 0.1650 & 0.1158 & 0.0609 \\
0.1635 & 0.1650 & 1.1665 & 0.1169 & 0.0615 \\
0.1148 & 0.1158 & 0.1169 & 1.1179 & 0.0620 \\
0.0604 & 0.0609 & 0.0615 & 0.0620 & 1.0625
\end{bmatrix}.
\]

In this case, we used a conditional masked autoregressive flow (MAF) \citep{Papamakarios2017} 
with standard normal base distribution as the parameterization of our proposals $q_1(x;y, \theta)$ and $q_2(x;y)$.
Here the normalizing flows consisted of 16 flow layers with single 1024 hidden units layer within each flow and we used tanh rather than relu activation functions as we found this made a significant difference in terms of training stability for the distribution $q_1$.
We did not find batch normalization to help the performance or stability significantly, and hence we have not used it.
We used the conditional MAF implementation from \href{http://github.com/ikostrikov/pytorch-flows}{http://github.com/ikostrikov/pytorch-flows}.

Training was done using importance sampling to generate the values of $\theta$ and $x$ as per~\eqref{eq:train-prop}
with \[
q^\prime(\theta, x) = p(\theta) \cdot \textsc{HalfNormal}(x; \mu=\theta, \sigma=\text{diag}(\Sigma_2)).
\]
We used a learning rate of $10^{-4}$ an the Adam optimizer.

The estimates of the ground truth values $\mu(y, \theta)$ were determined numerically using an SNIS estimator with $10^{10}$ samples
and the proposal 
$q(x; \theta) = \textsc{HalfNormal}(x; \mu=\theta, \sigma=\text{diag}(\Sigma_2))$.

\subsection{Planning Cancer Treatment}
\label{sec:cancer-model}

As explained in the main paper, this experiment revolves around an oncologist is trying to decide whether to administer a treatment to a cancer patient.
They have access to two noisy measurements of the tumor size, a simulator of tumor evolution, a model of the latent factors required for this simulator, and a loss function for administering the treatment given the final tumor size.
We note that this is problem for which the target function $f(x)$ does not have any changeable parameters (i.e. $\theta = \emptyset$).

The size of the tumor is measured at the time of admission $t\!=\!0$ and five days later ($t\!=\!5$), yielding observations
$c_0'$ and $c_5'$.
These are noisy measurements of the true sizes $c_0$ and $c_5$.
The loss function $\ell(c_{100})$ is based only on the size of the tumor after $t\!=\!100$ days of treatment.
The simulator for the development of the tumor takes the form of an ordinary differential equation (ODE) and is taken from~\citep{Hahnfeldt1999,Enderling2014,Rainforth2018nmc}.

The ODE itself is defined on two variables, the size of the tumor at time $t$, $c_t$, and corresponding carrying capacity, $K_t$, where we take $K_0\!=\!700$.
In addition to the initial tumor size $c_0$, the key parameter of the ODE, and the only one we model as varying across patients, is 
$\epsilon \!\in\! [0, 1]$, a coefficient determining the patient's response to the anti-tumor treatment.
The ODE now take the form
\begin{gather}
\label{eq:cancer-ode}
\hspace{-5pt}
\frac{\d c}{\d t} \!=\! - \lambda c \log \!\left( \frac{c}{K} \right) \!-\! \epsilon c
\quad\quad
\frac{\d K}{\d t} \!=\! \phi c \!-\! \psi K c^{2/3}
\end{gather}
where the values of the parameters 
$\phi \!=\! 5.85,\,
\psi \!=\! 0.00873,\,
\lambda \!=\! 0.1923$
are based on those recommended in \citet{Hahnfeldt1999}.
We use the notation
\begin{align}
c_t = \omega(K_0, c_0, \epsilon, t) 
\end{align}
to denote the deterministic process of running an ODE solver on~\eqref{eq:cancer-ode} with given inputs, up to time $t$,
and assume the following statistical model 
\begin{align}
\begin{split}
c_0 &\sim \textsc{Gamma}(k=25, \theta=20) \nonumber \\
\epsilon &\sim \textsc{Beta}(\alpha=5.0, \beta=10.0) \nonumber \\
c_t' &\sim \textsc{Gamma}\left(k=\frac{c_t^2}{10000}, \theta=\frac{c_t}{10000}\right).
\end{split}
\end{align}
To summarize and relate the model to the notation from Section~\ref{sec:amci}: 
$x\!= \! \{ c_0, \epsilon \}, y \!= \! \{ c_0' , c_1' \}$.
The function in this case is fixed to the loss function for administering the treatment given the final tumor size provided to us by the clinic
\begin{gather}
f(x) = \ell(\omega(700, c_0, \epsilon, t=100) ) \\
\ell(c) \!=\! \frac{1\!-\!2\! \times \!10^{-8}}{2} \! \left( \! \text{tanh}\! \left( \!- \frac{c\!-\!300}{150}\! \right) \!+\! 1 \! \right)\! + \!10^{-8}.
\end{gather}

\paragraph{Amortization}
In this case, the amortization is performed using parametric distributions as proposals: a Gamma distribution for $c_0$ and a Beta distribution for $\epsilon$, both parameterized by a multilayer perceptron with 16 layers with 5000 hidden units each.
Since we do not face an overwhelming mismatch between $f(x)$ and $p(x)$, unlike in the tail integral example,
the training was done by generating the values of $x$ from the prior $p(x)$ as per~\eqref{eq:train-prop-standard}.
We used a learning rate of $10^{-4}$ with the Adam optimizer.

Similarly to the case of five-dimensional tail integral example, 
the estimates serving as ground truth values $\mu(y)$ have been determined numerically using an SNIS estimator with $10^{9}$ samples
and the proposal set to the prior $q(x) = p(x)$.

\subsection{Mini-batching Procedure}

AMCI operates in a slightly unusual setting for neural network training because instead of having a fixed dataset, we are instead training on samples from our model $p(x,y)$.
The typical way to perform batch stochastic gradient optimization involves many epochs over the training dataset, stopping once the error increases on the validation set.
Each epoch is itself broken down into multiple iterations, wherein one takes a random mini-batch (subsample) 
from the dataset (without replacement) and updates the parameters based on a stochastic gradient step using these samples,
with the epoch finishing once the full dataset has been used.

However, there are different ways the training can proceed when we have the ability to generate an infinite amount of data 
from our model $p(x,y)$ and we now no longer fave the risk of overfitting.
There are two extremes approaches one could take.
The first one would be sampling two large but fixed-size datasets (training and validation)
before the time of training and then following the standard training procedure for the
finite datasets outlined above.
The other extreme would be to completely surrender the idea of dataset or epoch,
and sample each batch of data presented to the optimizer directly from $p(x,y)$.
In this case, we would not need a validation dataset as we would never be at risk 
of overfitting---we would finish the training once we are satisfied with the 
convergence of the loss value.

\citet{Paige2016} found that the method which empirically performed best in similar
amortized inference setting was one in the middle between the two extremes outlined above.
They suggest a method which decides when to sample new synthetic (training and validation) datasets,
based on performance on the validation data set.
They draw fixed-sized training and validation datasets and optimize the model using the standard finite data 
procedure on the training dataset until the validation error increases.
When that happens they sample new training and validation datasets and repeat the procedure.
This continues until empirical convergence of the loss value.
In practice, they allow a few missteps (steps of increasing value) for the validation loss
before they sample new synthetic datasets, 
and limit the maximum number of optimization epochs performed on a single dataset.

We use the above method throughout all of our experiments.
We allowed a maximum of 2 missteps w.r.t. the validation dataset and 
maximum of 30 epochs on a single dataset before sampling new datasets.

Note that the way training and validation datasets are generated is modified slightly when using the importance sampling
approach for generating $x$ and $\theta$ detailed in Section~\ref{sec:efficient-training}.
Whenever we use the objective in \eqref{eq:train-prop}, 
instead of sampling the training and validation datasets from the prior $p(x,y)$ we will sample
them from the distribution $q^\prime(\theta, x) \cdot p(y|x)$ 
where $q^\prime$ is a proposal chosen to be as close to $p(x)p(\theta)\!f(x;\theta)$ as possible.

We note that while training was robust to the number of missteps allowed, adopting the general scheme of \citet{Paige2016} was very important in achieving effective training: we initially tried generating every batch directly from the model $p(x,y)$ and 
we found that the proposals often converged to the local minimum of just sampling from the prior.

\section{Reusing samples}
\label{sec:reusing-samples}

The AMCI estimator in~\eqref{eq:estimator-full} requires taking $T=N+K+M$ samples, 
but only $N$, $K$, or $M$ are used to evaluate each of the individual estimators.
Given that, in practice, we do not have access to the perfectly optimal proposals,
it can sometimes be more efficient to reuse samples in the calculation of multiple components of the
expectation, particularly if the target function is cheap to evaluate relative to the proposal.
Care is required though to ensure that this is only done when a proposal remains valid (i.e. has finite
variance) for the different expectation.

To give a concrete example, in the case where $f(x;\theta)\ge0 \,\, \forall x,\theta$, such
that we can use a single proposal for the numerator as per~\eqref{eq:amci-mc-estimator}, we could use
the following estimator
\begin{align}
\label{eq:estimator-combination}
\mu(y,\theta)
\! \approx \!
\frac{\alpha \hat{E}_1(q_1) + (1-\alpha)\hat{E}_1(q_2)}
{\beta \hat{E}_2(q_1) + (1-\beta)\hat{E}_2(q_2)}
\end{align}
where $\hat{E}_i(q_j)$ indicates the estimate for $E_i$ using the samples from $q_j$.
The level of interpolation is set by parameters $\alpha, \beta$ which vary between 0 and 1.
If we had direct access to the optimal proposals, it would naturally be
preferable to set $\alpha\! =\! 1$ and $\beta\!=\! 0$, leading to a zero-variance
estimator.
However, for imperfect proposals, the optimal values vary slightly from this 
(see \hyperref[sec:opt-vals]{Appendix~\ref*{sec:opt-vals}}).

\begin{figure*}[t!]
	\centering
	\includegraphics[width=0.86\textwidth]{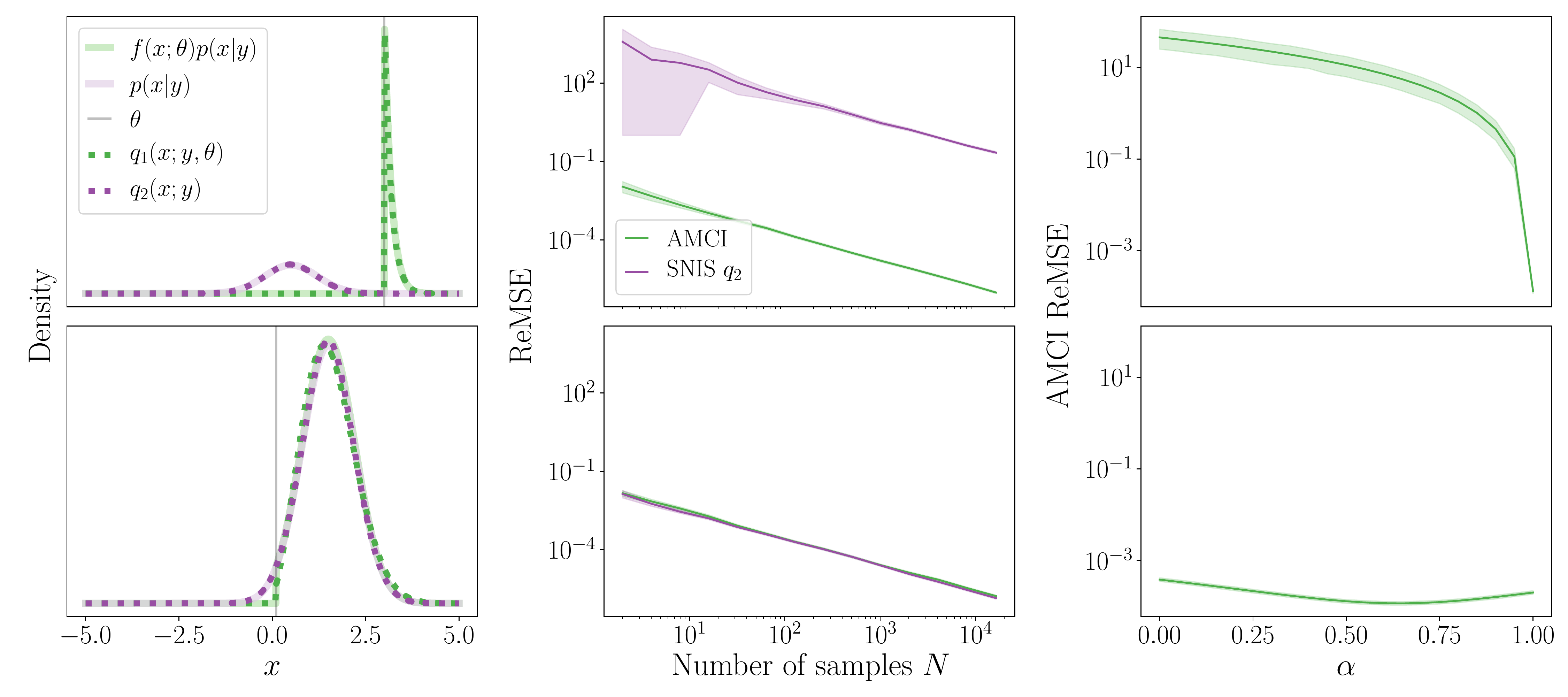}
	\vspace{-12pt}
	\caption{
		Extension of Figure~\ref{fig:tail-1d-figure}.
		Column three presents the effects of reusing samples by varying the parameter $\alpha$ in~\eqref{eq:estimator-combination} ($\beta=0$, number of samples is fixed to $N\!=\!M\!=\!64$), where we see that this sample re-usage provides small gains for the low mismatch case, but no gains in the high mismatch case.
		Uncertainty bands in columns two and three are estimated over a 1000 runs and are very small.
	}\vspace{-12pt}
	\label{fig:tail-1d-figure-extended}
\end{figure*}

In relation to our discussion in Section~\ref{sec:discussion}, 
the third column of Figure~\ref{fig:tail-1d-figure-extended}
shows how when $f(x;\theta)p(x,y)$ and $p(x,y)$ are closely matched
we can decrease the error of our AMCI estimator by reusing samples
through setting $\alpha < 1$.

Note that while it is possible to set $\beta>0$ for negligible extra computational cost
as $\hat{E}_2(q_1)$ depends only on weights needed for calculating $\hat{E}_1(q_1)$,
setting $\alpha<1$ requires additional evaluations of the target function and so will
likely only be beneficial when this is cheap relative to sampling from or evaluating the proposal.

\subsection{Derivation of the optimal parameter values for $\alpha$ and $\beta$}
\label{sec:opt-vals}

In this section, we derive the optimal values of $\alpha$ and $\beta$ in terms of minimizing the mean squared error (MSE) 
of the estimator in \eqref{eq:estimator-combination}. 
We assume that we are allocated a total sample budget of $T$ samples, such
that $M=T-N$.

Let the true values of the expectations in the numerator and denominator be 
denoted as $E_1$ and $E_2$, respectively.
We also define the following shorthands for the unbiased importance sampling estimators with respect to proposals $q_1$ and $q_2$ in \eqref{eq:estimator-combination}
$a_1 = \frac{1}{N} \sum_n^N \frac{f(x_n; \theta) p(x_n, y) }{q_1(x_n;y, \theta)}$,
$b_1 = \frac{1}{M} \sum_m^M \frac{f(x_m^*; \theta) p(x_m^*, y)}{q_2(x_m^*;y)}$,
$a_2 = \frac{1}{N} \sum_n^N \frac{p(x_n, y)}{q_1(x_n;y, \theta)}$,
$b_2 = \frac{1}{M} \sum_m^M \frac{p(x_m^*, y)}{q_2(x_m^*;y)}$,
where $x_n \sim q_1(x;y, \theta)$ and $x_m^* \sim q_2(x;y)$.

We start by considering the estimator according to~\eqref{eq:estimator-combination}
\begin{align}
	\mu :=
	\frac{E_1 }{E_2}
  \approx \hat{\mu} :=
	\frac{\hat{E}_1 }{\hat{E}_2} :=& 
	\frac{\alpha a_1 + (1 - \alpha) b_1}{\beta a_2 + (1 - \beta) b_2}.
\end{align}	
Using the central limit theorem separately for $\hat{E}_1$ and $\hat{E}_2$, then we thus have, as $N,M \to \infty$,
\begin{align}
\hat{\mu}\to&~ \frac{E_1 + \sigma_1 \xi_1}{E_2 + \sigma_2 \xi_2},
  \label{eq:asy-est}
\end{align}  
where $\xi_1, \xi_2 \sim \mathcal{N}(0,1)$ are correlated standard normal random variables and
		$\sigma_1$ and  $\sigma_2$ are the standard deviation of the 
		estimators for the numerator and the denominator, respectively.
Specifically we have
\begin{align}
	  \sigma_1^2 
	=& \Var[\alpha a_1 + (1-\alpha) b_1] \nonumber \\
	=& \alpha^2 \Var_{q_1}[a_1] + (1-\alpha)^2 \Var_{q_2}[b_1], \nonumber\\
	\intertext{which by the weak law of large numbers}
	=& \frac{\alpha^2}{N} \vf + \frac{(1-\alpha)^2}{M} \vfs \\
	\intertext{where $w_1=p(x_1,y)/q_1(x_1;y, \theta)$, $w_1^*=p(x_1^*,y)/q_2(x_1^*;y)$, $x_1 \sim q_1(x;y, \theta)$, and  $x_1^* \sim q_2(x;y)$.  Analogously,}
	\sigma_2^2 =& \frac{\beta^2}{N} \Var_{q_1}[w_1] + \frac{(1-\beta)^2}{M} \Var_{q_2}[w^*_1].	
\end{align}  
		
Now going back to~\eqref{eq:asy-est} and using Taylor's Theorem on $1/ \left(E_2 + \sigma_2 \xi_2 \right)$ about $1/E_2$ gives
\begin{align}
  \hat{\mu}=& \frac{E_1 + \sigma_1 \xi_1}{E_2} 
  \left(1 - \frac{\sigma_2 \xi_2}{E_2}\right) + O(\epsilon) \nonumber\\
  =& \frac{E_1}{E_2} + \frac{\sigma_1 \xi_1}{E_2} - \frac{E_1 \sigma_2 \xi_2}{E_2^2} - \frac{\sigma_1 \sigma_2 \xi_1 \xi_2}{E_2^2} + O(\epsilon) \nonumber\\
  \intertext{where $O(\epsilon)$ represents asymptotically dominated terms.  
  	Note here
  	the importance of using Taylor's theorem, instead of just a Taylor expansion,
  	to confirm that these terms are indeed asymptotically dominated.
  	We can further drop the 
  		$\sigma_1 \sigma_2 \xi_1 \xi_2 / E_2^2$ term as this will be of
  		order $O(1/\sqrt{MN})$ and will thus be asymptotically dominated, giving
  	}
	  =& \frac{E_1}{E_2} + \frac{\sigma_1 \xi_1}{E_2} - \frac{E_1 \sigma_2 \xi_2}{E_2^2} + O(\epsilon).
\end{align}
	  
To calculate the MSE of $\hat{\mu}$, we start with the standard bias variance
decomposition
\begin{align}
\mathbb{E} \left[\left(\hat{\mu}-\frac{E_1}{E_2}\right)^2\right]
= \Var \left[\hat{\mu}\right]
+\left(\mathbb{E} \left[\hat{\mu}-\frac{E_1}{E_2}\right]\right)^2.
\end{align}
Considering first the bias squared term, we see that this depends only
on the higher order terms $O(\epsilon)$, while the variance does not.
It straightforwardly follows that the variance term will be asymptotically
dominant, so we see that optimizing for the variance is asymptotically
equivalent to optimizing for the MSE.

Now using the standard relationship $\operatorname{Var} [X \! + \! Y]\!=\!\operatorname{Var} [X]\!+\! \operatorname{Var} [Y] \!+ \! 2 \operatorname{Cov} [X,Y]$ yields
\begin{align}
  \Var[\hat{\mu}] 
  =& \,\, \Var\left[\frac{E_1}{E_2}\right] + \Var\left[\frac{\sigma_1 \xi_1}{E_2}\right] + \Var\left[\frac{E_1 \sigma_2 \xi_2}{E_2^2}\right] 
  + 2 \operatorname{Cov} \left[\frac{\sigma_1 \xi_1}{E_2}, -\frac{E_1 \sigma_2 \xi_2}{E_2^2}\right] + O(\epsilon) \displaybreak[0] \nonumber\\
  \approx& \,\, 0 + \frac{\sigma_1^2}{E_2^2} + \frac{E_1^2 \sigma_2^2}{E_2^4}
  - 2 \frac{E_1 \sigma_1 \sigma_2}{E_2^3} \operatorname{Cov}[\xi_1, \xi_2] 
	\nonumber\\
	=& \,\, \frac{1}{E_2^2} \left( \sigma_1^2 + \sigma_2^2 \mu^2 - 2 \mu \sigma_1 \sigma_2  \text{Corr}\left[\xi_1, \xi_2 \right] \right)
	\label{eq:amci-mu-variance}
  \displaybreak[0]\\
	\intertext{since $\Var[\xi_1]=\Var[\xi_2]=1 \implies \operatorname{Cov}[\xi_1, \xi_2]=\text{Corr}\left[\xi_1, \xi_2 \right]$,}
  =& \,\, \frac{\alpha^2}{N E_2^2} \vf + \frac{(1-\alpha)^2}{M E_2^2} \vfs + \frac{E_1^2 \beta^2}{N E_2^4} \Var_{q_1}[w_1] + \frac{E_1^2 (1-\beta)^2}{M E_2^4} \Var_{q_2}[w^*_1]\nonumber\\
  &- 2 \frac{E_1}{E_2^3}  \text{Corr}[\xi_1, \xi_2]
  \left(\frac{\alpha^2}{N} \vf + \frac{(1-\alpha)^2}{M} \vfs\right) \left(\frac{\beta^2}{N} \Var_{q_1}[w_1] + \frac{(1-\beta)^2}{M} \Var_{q_2}[w^*_1]\right) \nonumber
\end{align}
To assist in the subsequent analysis, we assume that there is no correlation,
$\text{Corr}[\xi_1, \xi_2]=0$.
Though this assumption is unlikely to be exactly true, 
there are two reasons we believe it is reasonable.  Firstly, because we expect
to set $\alpha\approx 1$ and $\beta \approx 0$, the correlation should 
generally be small in practice as the two estimators rely predominantly on
independent sets of samples.  Secondly, we believe this is generally a relatively
conservative assumption: if one were to presume a particular correlation,
there are adversarial cases with the opposite correlation where this assumption
is damaging.  

Given this assumption it is now straightforward to optimize for $\alpha$ and
$\beta$ by finding where the gradient is zero as follows
\begin{align}
  \grad_\alpha (\Var[\hat{\mu}] E_2^2)
  &= \frac{2 \alpha \vf}{N} - \frac{2 (1-\alpha) \vfs}{T-N}
  = 0 \nonumber\\
\Rightarrow  \alpha^* 
  &= N \cdot \left( (T-N) \frac{\vf}{\vfs} + N \right)^{-1}
  \intertext{noting that}
  \grad^2_\alpha (\Var[\hat{\mu}] E_2^2) 
	&= \frac{\vf}{N} + \frac{\vfs}{T-N} 
  > 0 \nonumber
\intertext{and hence it's a local minimum. Analogously}
  \beta^* 
  &= N \cdot \left( (T-N) \frac{\Var_{q_1}[w_1]}{\Var_{q_2}[w^*_1]} + N \right)^{-1}.
\end{align}
We note that it is possible to estimate all the required variances here using
previous samples.  It should therefore be possible to adaptively set
$\alpha$ and $\beta$ by using these equations along with empirical
estimates for these variances.

\end{document}